\documentclass{article} % For LaTeX2e
\usepackage[final]{colm2026_conference}
\usepackage{microtype}
\usepackage{hyperref}
\usepackage{url}
\usepackage{booktabs}
\usepackage{amsmath}
\usepackage{algorithm}
\usepackage{algorithmic}
\usepackage{multirow}
\usepackage{enumitem}
\usepackage{graphicx}
\usepackage{wrapfig}
\usepackage{subcaption}
\setlist[itemize]{nosep}
\usepackage{titlesec}

\usepackage{float}
\titlespacing*{\paragraph}{0pt}{2pt}{2pt}
\usepackage{lineno}

\definecolor{darkblue}{rgb}{0, 0, 0.5}
\hypersetup{colorlinks=true, citecolor=darkblue, linkcolor=darkblue, urlcolor=darkblue}

\title{RefusalGuard: Geometry-Preserving Fine-Tuning for Safety in LLMs}

\author{
Sadia Asif , Mohammad Mohammadi Amiri \\
Department of Computer Science \\
Rensselaer Polytechnic Institute \\
Troy, NY, USA \\
\texttt{\{asifs, mamiri\}@rpi.edu}
}

\begin{document}

\ifcolmsubmission
\linenumbers
\fi

\maketitle

\begin{abstract}
Fine-tuning safety-aligned language models for downstream tasks often leads to substantial degradation of refusal behavior, making models vulnerable to adversarial misuse. While prior work has shown that safety-relevant features are encoded in structured representations within the model’s activation space, how these representations change during fine-tuning and why alignment degrades remains poorly understood. In this work, we investigate the representation-level mechanisms underlying alignment degradation. Our analysis shows that standard fine-tuning induces systematic drift in safety-relevant representations, distorts their geometric structure, and introduces interference between task optimization and safety features. These effects collectively lead to increased harmful compliance. Motivated by these findings, we introduce \textsc{RefusalGuard}, a representation-level fine-tuning framework that preserves safety-relevant structure during model adaptation. Our approach constrains updates in hidden representation space, ensuring that safety-mediating components remain stable while allowing task-specific learning in complementary directions. We evaluate \textsc{RefusalGuard} across multiple model families, including LLaMA, Gemma, and Qwen, on adversarial safety benchmarks such as AdvBench, DirectHarm4, and JailbreakBench, as well as downstream utility tasks.  Our approach achieves attack success rates comparable to base safety-aligned models while maintaining competitive task performance, significantly outperforming baselines.

\end{abstract}
\section{Introduction}

Large language models (LLMs) are increasingly deployed through post-training customization, where a safety-aligned base model is further fine-tuned for domain-specific tasks, user preferences, or application-specific behaviors \citep{ouyang2022training,touvron2023llama2,zhang2024scalingfinetuning}. In practice, this adaptation step is often essential: models must be specialized to downstream distributions while retaining the instruction-following and refusal behaviors established during alignment. However, recent work has shown that this assumption is fragile. Even when the fine-tuning data is benign, or when the model is exposed to only a very small number of harmful examples, aligned models can exhibit a substantial loss of safety, becoming more likely to comply with harmful or adversarially phrased requests \citep{qi2023finetuning, fraser2025finetuning, hsiung2025guardrails}. This creates a practical and increasingly important failure mode: a model that is safe at release time may become unsafe after  standard post-alignment adaptation.

This phenomenon raises a central question: \emph{why does fine-tuning degrade refusal behavior in aligned LLMs?} Existing work has largely studied this problem from the level of prompts, datasets, or attack outcomes. These studies establish that alignment can be brittle under both malicious and benign fine-tuning, but they provide limited visibility into the internal mechanisms by which safety is lost \citep{wehner2025taxonomy, xie2025deeprefusal, das2025alignguard}. As a result, current mitigation strategies are often heuristic, focusing on prompt design, additional safety examples, or adversarial training (For details see Appendix \ref{sec:related-work}), without a clear account of what safety-relevant internal structure should be preserved during adaptation.

At the same time, recent interpretability work suggests that refusal behavior is not arbitrary, but mediated by structured representations in activation space. Prior studies show that safety-relevant concepts can often be localized in hidden states and manipulated through representation-level interventions \citep{wu2024reft,zhang2025rational}. In particular, recent work on refusal geometry argues that refusal is associated not with a single direction or an unstructured collection of activations, but with coherent directions or low-dimensional geometric structure in the residual stream \citep{arditi2024refusal}. These findings provide an important clue: if refusal is mediated by identifiable representational structure, then alignment degradation under fine-tuning may arise because this structure is displaced, distorted, or interfered with during optimization.

Despite this promise, an important gap remains. Prior interpretability work has mainly focused on identifying refusal-related features and analyzing their causal role under intervention \citep{siu2025cosmic,marshall2024refusal, mazeika2024harmbench}. It does not explain how these safety-relevant structures evolve during downstream fine-tuning, nor does it provide a training method that explicitly preserves them. Conversely, prior work on safe fine-tuning and jailbreak robustness proposes practical defenses \citep{andriushchenko2024jailbreaking}, but generally does not operate at the level of internal safety geometry \citep{das2025alignguard}. Thus, we currently lack a principled representation-level account of alignment degradation and a corresponding fine-tuning strategy designed to prevent it.

In this work, we study alignment degradation through the lens of representation geometry. We analyze how safety-relevant representations change during standard fine-tuning and show that degradation is accompanied by three consistent effects: (i) directional drift in refusal-mediating representations, (ii) distortion of their geometric structure, and (iii) growing interference between task-driven updates and safety-relevant components. These changes jointly weaken refusal behavior and increase harmful compliance. Our analysis suggests that the problem is not merely that fine-tuning adds new task capabilities; rather, it can also overwrite or destabilize the internal structures that support safe refusal.

Motivated by this observation, we introduce \textsc{RefusalGuard}, a representation-level fine-tuning framework for preserving safety-relevant structure during adaptation. The key idea is to constrain hidden-state updates so that safety-mediating components remain stable while allowing learning to proceed in complementary directions that support downstream utility. Rather than relying solely on additional safety supervision or prompt-time interventions, \textsc{RefusalGuard} directly targets the internal representational changes associated with alignment degradation. This yields a more mechanistically grounded approach to maintaining safety under fine-tuning.

We evaluate \textsc{RefusalGuard} across multiple model families, including LLaMA \citep{touvron2023llama}, Gemma \citep{gemmateam2024gemma}, and Qwen \citep{yang2025qwen3}, on adversarial safety benchmarks such as AdvBench \citep{zou2023universaladvbench}, DirectHarm4 \citep{lyu2024prompttemplatesdirectharm4}, and JailbreakBench \citep{chao2024jailbreakbench}, together with downstream utility tasks such as GSM8k \citep{cobbe2021traininggsm8k}, OpenOrca \citep{openrca2025llm}, and ARC \citep{moskvichev2023conceptarc}. Empirically, \textsc{RefusalGuard} substantially improves safety preservation during fine-tuning, achieving attack success rates comparable to the original safety-aligned models while maintaining competitive downstream performance. These results indicate that preserving safety-relevant geometry is a viable and effective principle for robust model adaptation.

Our contributions can be summarized as follows:
% \vspace{-6pt}
\begin{itemize}[itemsep=0pt]
    \item We provide a representation-level analysis of alignment degradation under fine-tuning, showing that refusal behavior deteriorates through systematic drift, geometric distortion, and optimization interference in safety-relevant representations.
    \item We propose \textsc{RefusalGuard}, a geometry-preserving fine-tuning framework that stabilizes safety-mediating components in hidden representation space while preserving task-specific adaptation.
    \item We demonstrate across several model families and safety benchmarks that representation-level preservation can substantially improve post-fine-tuning safety without sacrificing competitive downstream utility.
\end{itemize}
\vspace{-6pt}

\section{Mechanistic Analysis of Refusal Geometry Under Fine-Tuning}
\label{sec:mechanistic-analysis}

In this section, we use refusal geometry as a mechanistic lens to study \emph{how} fine-tuning weakens refusal behavior. Our analysis builds on recent work showing that refusal is not governed by a single isolated feature, but by a low-dimensional geometric structure in the residual stream. In particular, prior work demonstrates that refusal-mediating directions can form multi-dimensional \emph{concept cones}, and that orthogonality alone does not imply mechanistic independence under intervention \citep{arditi2024refusal,wollschlager2025geometry}. We use these findings as the starting point for analyzing what downstream fine-tuning does to safety-relevant internal structure.

\subsection{Refusal Cones as a Reference Geometry}
\label{subsec:reference-geometry}

Let $f_{\theta_0}$ denote the original safety-aligned model before downstream fine-tuning. Following prior work on refusal geometry, we identify a low-dimensional refusal cone in the \emph{hidden representation space at layer $\ell$}. Concretely, let
\begin{equation}
\boldsymbol{h}^{(\ell)}_\theta(x) \in \mathbb{R}^d
\end{equation}
denote the $d$-dimensional hidden representations at layer $\ell$ for input $x$.

We extract a $k$-dimensional ($k \ll d$) refusal subspace with orthonormal basis
\begin{equation}
\boldsymbol{B}_0 = [\boldsymbol{b}_1,\dots,\boldsymbol{b}_k] \in \mathbb{R}^{d \times k}, 
\quad 
\end{equation}
where $\boldsymbol{B}_0$ spans the refusal-relevant directions in representation space. The corresponding refusal cone is defined as
\begin{equation}
\mathcal{C}_0 = \left\{ \sum_{j=1}^{k} \alpha_j \boldsymbol{b}_j \;:\; \alpha_j \geq 0 \right\}.
\end{equation}

Here, $d$ is the residual-stream dimension and $k$ is the cone dimension. Intuitively, $\mathcal{C}_0$ captures a region of activation space whose directions mediate refusal in the original aligned model. For an input prompt $x$, let $\boldsymbol{h}_\theta^{(\ell)}(x) \in \mathbb{R}^d$ denote the hidden representations at layer $\ell$ under parameters $\theta$. To study how refusal-relevant structure changes during fine-tuning, we analyze the evolution of harmful-input representations relative to the reference cone $\mathcal{C}_0$, and when needed, compare $\mathcal{C}_0$ to a cone $\mathcal{C}_t$ re-estimated from a fine-tuned checkpoint $\theta_t$ using the same extraction procedure.

This gives two complementary perspectives: (i) \textbf{Anchored analysis:} track how activations of the fine-tuned model move relative to the refusal geometry of the original aligned model. (ii) \textbf{Comparative analysis:} re-estimate refusal geometry after fine-tuning and compare the resulting cone to the original one.
% \end{itemize}
The first isolates whether safety-relevant components are preserved; the second reveals whether fine-tuning changes the underlying geometry itself.

\subsection{What Changes Under Fine-Tuning?}
\label{subsec:what-changes}

We study three mechanisms that are consistent with the safety degradation observed after downstream fine-tuning.

\paragraph{(i) Directional drift.}
If refusal behavior depends on a structured region of activation space, then fine-tuning may degrade safety by moving harmful-input representations away from that region. To quantify this effect, we project the harmful activation onto the reference cone basis:
\begin{equation}
\boldsymbol{z}_{\theta}^{(\ell)}(x) = \boldsymbol{B}_0^\top \boldsymbol{h}_{\theta}^{(\ell)}(x) \in \mathbb{R}^k.
\end{equation}
The quantity $\|\boldsymbol{z}_{\theta}^{(\ell)}(x)\|_2$ measures how strongly the activation aligns with the base refusal subspace, while the coordinates of $\boldsymbol{z}_{\theta}^{(\ell)}(x)$ indicate how this alignment is distributed across the cone basis. A systematic reduction in this projected magnitude after fine-tuning indicates that harmful prompts are no longer activating the safety-relevant structure as strongly as in the base aligned model.

To isolate directional change independent of norm, we also track the cosine similarity between the activation and its cone projection:
\begin{equation}
\mathrm{Align}_{\theta}^{(\ell)}(x)
=
\frac{\langle \boldsymbol{h}_{\theta}^{(\ell)}(x),\, \boldsymbol{B}_0 \boldsymbol{B}_0^\top \boldsymbol{h}_{\theta}^{(\ell)}(x)\rangle}
{\|\boldsymbol{h}_{\theta}^{(\ell)}(x)\|_2 \, \|\boldsymbol{B}_0 \boldsymbol{B}_0^\top \boldsymbol{h}_{\theta}^{(\ell)}(x)\|_2}.
\end{equation}
A drop in this quantity indicates directional drift away from the refusal-mediating region.

\paragraph{(ii) Geometric distortion.}
A second possibility is that fine-tuning does not simply rotate activations away from refusal geometry, but changes the geometry itself. To test this, we re-estimate a refusal cone at fine-tuning checkpoint $\theta_t$, with orthonormal basis $\boldsymbol{B}_t = [\boldsymbol{b}_{t,1}, \dots, \boldsymbol{b}_{t,k}] \in \mathbb{R}^{d \times k}$, and compare its span to that of $\boldsymbol{B}_0$. We quantify cone drift through principal-angle or subspace-overlap measures between $\mathrm{span}(\boldsymbol{B}_0)$ and $\mathrm{span}(\boldsymbol{B}_t)$. One convenient summary is
\begin{equation}
\mathrm{Drift}(\boldsymbol{B}_0,\boldsymbol{B}_t)
=
1 - \frac{1}{k}\|\boldsymbol{B}_0^\top \boldsymbol{B}_t\|_*,
\end{equation}
where $\|\cdot\|_*$ denotes the nuclear norm. This value is small when the two subspaces remain close and increases as the fine-tuned refusal geometry rotates away from the original one.

Subspace drift alone, however, does not fully characterize geometric degradation. Since refusal is modeled as a cone rather than an unconstrained subspace, the \emph{relative organization} of directions also matters. We therefore examine how the coordinates of harmful activations within the cone change over training. In particular, if fine-tuning compresses harmful activations toward a few unstable directions or pushes them toward the cone boundary, then refusal becomes more brittle: small downstream perturbations can more easily move the representation outside the refusal region. We refer to this effect as \emph{cone distortion}. Operationally, it appears as reduced and less stable support across cone coordinates, together with increased concentration near a subset of directions.

\paragraph{(iii) Task-safety interference.}
The previous two effects describe what happens in activation space; the third addresses \emph{why} these changes arise during optimization. Let
$\mathcal{L}_{\mathrm{task}}$ denote the downstream fine-tuning objective. For a harmful prompt $x'$, the task gradient induces a first-order change in the hidden state
\begin{equation}
\Delta \boldsymbol{h}^{(\ell)}(x')
\approx
-\eta \nabla_{\boldsymbol{h}^{(\ell)}} \mathcal{L}_{\mathrm{task}}(x'),
\end{equation}
where $\eta$ is the learning rate and $\nabla_{\boldsymbol{h}^{(\ell)}} \mathcal{L}_{\mathrm{task}}(x')$ denotes the gradient of the task loss with respect to the layer-$\ell$ hidden representations. We measure whether this update interferes with refusal geometry by decomposing it into cone-parallel and cone-orthogonal components:
\begin{equation}
\Delta \boldsymbol{h}^{(\ell)}_{\text{Par}}
=
\boldsymbol{B}_0 \boldsymbol{B}_0^\top \Delta \boldsymbol{h}^{(\ell)},
\qquad
\Delta \boldsymbol{h}^{(\ell)}_{\text{Oth}}
=
\Delta \boldsymbol{h}^{(\ell)} - \Delta \boldsymbol{h}^{(\ell)}_{\text{Par}}.
\end{equation}
If downstream fine-tuning were safety-compatible, most task-driven movement should occur in directions complementary to the refusal cone. In contrast, a large cone-parallel component, especially one that opposes the sign structure associated with refusal, indicates direct interference between task optimization and safety-mediating features. We summarize this effect using a normalized interference score
\begin{equation}
\mathrm{Interf}^{(\ell)}(x')
=
\frac{\|\Delta \boldsymbol{h}^{(\ell)}_{\text{Par}}(x')\|_2}
{\|\Delta \boldsymbol{h}^{(\ell)}(x')\|_2}.
\end{equation}
Higher values indicate that task optimization is acting substantially along safety-relevant directions rather than around them.

\subsection{A Mechanistic Picture of Alignment Degradation}
\label{subsec:mechanistic-picture}

Figure~\ref{fig:mechanistic_harmful_10}, together with additional results in Appendix~\ref{app:ablation}, reveals how refusal-relevant structure evolves and degrades under targeted harmful fine-tuning. Even when the model is exposed to only 10 harmful examples, we observe rapid and systematic degradation across all mechanistic dimensions. Alignment to the reference cone decreases steadily over training, indicating that harmful inputs increasingly fail to activate safety-relevant directions. This effect is mirrored by a consistent drop in projected refusal magnitude, showing that the strength of activation within the refusal subspace is also diminished. At the same time, geometric instability becomes pronounced. Cone drift increases monotonically, demonstrating that the underlying refusal geometry itself is being displaced rather than merely underutilized. This is accompanied by a sharp rise in task-safety interference, indicating that gradient updates are not orthogonal to safety-relevant directions but instead actively modify them. Notably, these geometric changes closely track the increase in attack success rate, suggesting a tight coupling between internal representation drift and externally observable safety failures. 

Taken together, these trends provide a consistent mechanistic explanation for alignment degradation. This analysis leads to an important conclusion for the remainder of the paper. The safety loss induced by fine-tuning is not adequately explained as generic catastrophic forgetting alone. Rather, it reflects a specific geometric failure: the internal structures that mediate refusal are displaced and destabilized by downstream optimization. This observation motivates our method in the next section. If safety-relevant geometry is what is being damaged, then preserving that geometry during fine-tuning should provide a principled way to maintain refusal behavior while still allowing task adaptation.

\begin{figure*}[t]
\vspace{-6pt}
\centering
\includegraphics[width=\textwidth,height=0.32\textheight,keepaspectratio]{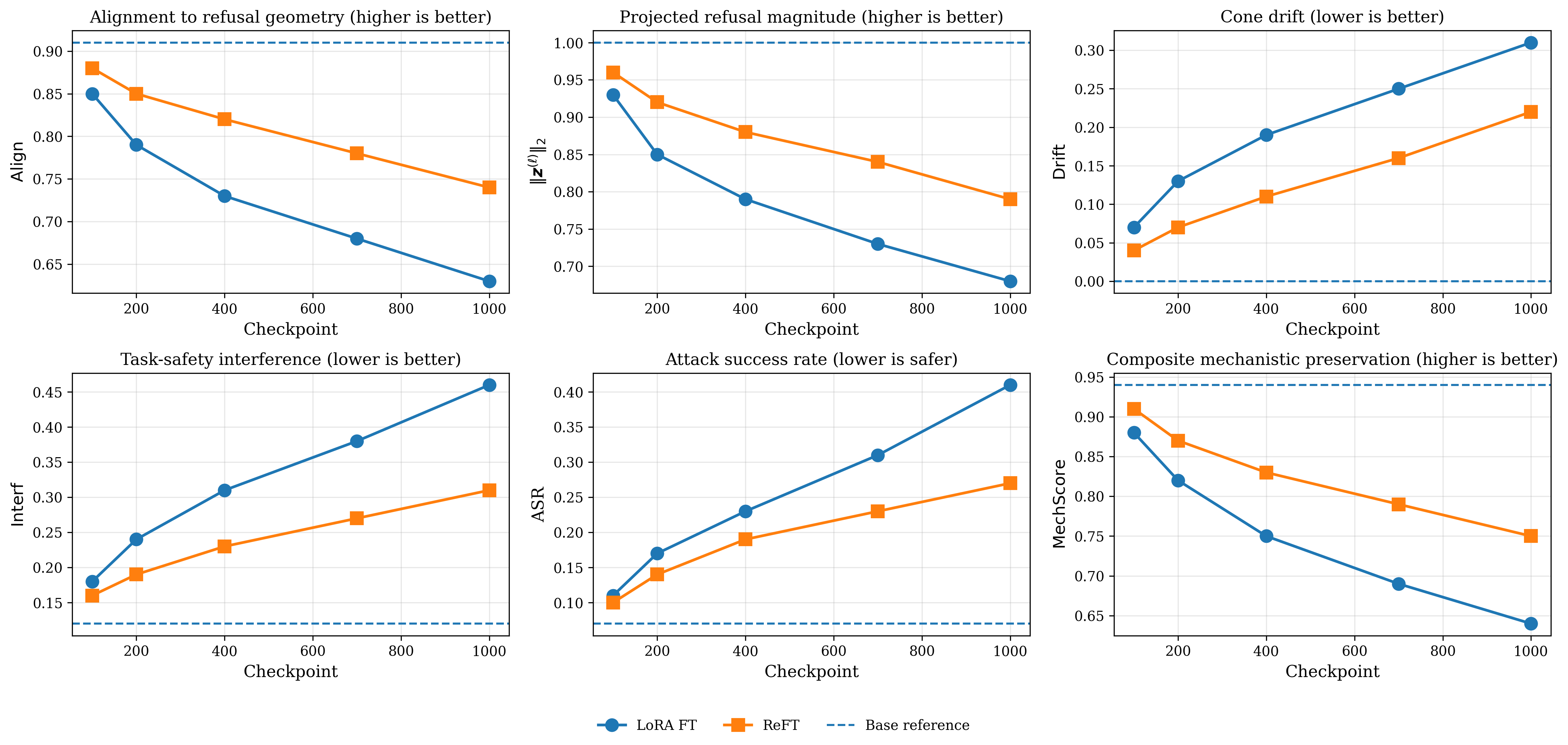}
\vspace{-10pt}
\caption{
Mechanistic degradation under harmful fine-tuning on \textbf{LLaMA-3.1-8B-Instruct} using \textbf{10 synthetic harmful examples}.
Both LoRA and Reft rapidly degrade refusal structure with respect to base aligned model: alignment and magnitude decrease, while drift, interference, and ASR increase.
}
\vspace{-10pt}
\label{fig:mechanistic_harmful_10}
\end{figure*}
\section{Methodology}
\label{sec:method}

The analysis in Section~\ref{sec:mechanistic-analysis} suggests that safety degradation under standard fine-tuning is driven by movement of hidden representations along refusal-relevant directions. This motivates a simple design principle: adapt the model through representation-level interventions, but discourage those interventions from modifying the portion of the representations that lies inside the refusal geometry. To implement this idea, we build on Representation Finetuning (ReFT) \citep{wu2024reft}, which learns interventions on hidden states while keeping the underlying language model frozen. Our method, \textsc{RefusalGuard}, follows the same overall paradigm, but augments it with a soft geometric constraint that preserves refusal-relevant structure during downstream adaptation.
\subsection{Preliminaries and Fine-Tuning Setup}
\label{subsec:method-preliminaries}

Let $f_{\theta_0}$ denote a frozen safety-aligned language model. Given an input sequence $x = (x_1,\dots,x_n)$, let $\boldsymbol{h}_p^{(\ell)}(x) \in \mathbb{R}^{d_\ell}$ denote the hidden representations at token position $p$ and layer $\ell$, where $d_\ell$ is the residual-stream dimension. For notational simplicity, we omit explicit dependence on $\theta_0$. We do not update the base model parameters $\theta_0$. Instead, we learn a set of intervention modules that act on hidden states during the forward pass. An intervention is specified by a layer $\ell$, a set of token positions $\mathcal{P}_\ell(x) \subseteq \{1,\dots,n\}$, and a learned map $\Phi_{\phi}^{(\ell)} : \mathbb{R}^{d_\ell} \to \mathbb{R}^{d_\ell}$, where $\phi$ denotes the collection of all learnable parameters of the intervention modules. For every edited position $p \in \mathcal{P}_\ell(x)$, the hidden state is replaced by $\tilde{\boldsymbol{h}}_p^{(\ell)} = \Phi_{\phi}^{(\ell)}(\boldsymbol{h}_p^{(\ell)})$, and the forward pass proceeds using the intervened representation.
\paragraph{Refusal geometry.}
For each intervention layer $\ell$, we extract from the aligned base model a $k_\ell$-dimensional refusal cone. To distinguish this layer-specific object from the reference cone introduced in Section~\ref{sec:mechanistic-analysis}, we denote its orthonormal basis by
\begin{equation}
\boldsymbol{B}_{\mathrm{ref}}^{(\ell)} = [\boldsymbol{b}^{(\ell)}_1,\dots,\boldsymbol{b}^{(\ell)}_{k_\ell}] \in \mathbb{R}^{d_\ell \times k_\ell},
\qquad
(\boldsymbol{B}_{\mathrm{ref}}^{(\ell)})^\top \boldsymbol{B}_{\mathrm{ref}}^{(\ell)} = \boldsymbol{I}.
\end{equation}

The span of $\boldsymbol{B}_{\mathrm{ref}}^{(\ell)}$ defines the refusal-relevant subspace at layer $\ell$. The associated projection matrix
\begin{equation}
\boldsymbol{P}_{\mathrm{ref}}^{(\ell)} = \boldsymbol{B}_{\mathrm{ref}}^{(\ell)} (\boldsymbol{B}_{\mathrm{ref}}^{(\ell)})^\top \in \mathbb{R}^{d_\ell \times d_\ell}
\end{equation}
is the orthogonal projection onto this subspace. The complementary projection is
\begin{equation}
\boldsymbol{P}_{cmp}^{(\ell)} = \boldsymbol{I} - \boldsymbol{P}_{\mathrm{ref}}^{(\ell)}.
\end{equation}

Thus, any hidden state at layer $\ell$ decomposes as
\begin{equation}
\boldsymbol{h}_p^{(\ell)}
=
\boldsymbol{P}_{\mathrm{ref}}^{(\ell)} \boldsymbol{h}_p^{(\ell)}
+
\boldsymbol{P}_{cmp}^{(\ell)} \boldsymbol{h}_p^{(\ell)}.
\label{eq:rep-decomp}
\end{equation}
The first term captures the safety-relevant component identified by the refusal geometry, while the second captures directions available for task adaptation.

\subsection{ Representation Intervention}
\label{subsec:method-intervention}

We parameterize each intervention using a standard low-rank ReFT map. Concretely, for a hidden state
$\boldsymbol{h} \in \mathbb{R}^{d_\ell}$, we define
\begin{equation}
\Phi_\phi^{(\ell)}(\boldsymbol{h})
=
\boldsymbol{h} + \Delta_\phi^{(\ell)}(\boldsymbol{h}),
\label{eq:intervention-general}
\end{equation}
where $\Delta_\phi^{(\ell)}(\boldsymbol{h})$ is a learned low-rank edit. Following the LoReFT parameterization \citep{wu2024reft}, one convenient choice is
\begin{equation}
\Delta_\phi^{(\ell)}(\boldsymbol{h})
=
({\boldsymbol{R}^{(\ell)}})^\top
\Big(
\boldsymbol{W}^{(\ell)} \boldsymbol{h} + \boldsymbol{b}^{(\ell)} - \boldsymbol{R}^{(\ell)} \boldsymbol{h}
\Big),
\label{eq:loreft-edit}
\end{equation}
with $\boldsymbol{R}^{(\ell)} \in \mathbb{R}^{r_\ell \times d_\ell}, \boldsymbol{W}^{(\ell)} \in \mathbb{R}^{r_\ell \times d_\ell}, \boldsymbol{b}^{(\ell)} \in \mathbb{R}^{r_\ell}$,
where $r_\ell \ll d_\ell$ is the intervention rank. Intuitively, it computes a low-dimensional edit and lifts it back to the original hidden-state space. Equations~\eqref{eq:intervention-general}--\eqref{eq:loreft-edit} define the task-adaptation mechanism. By themselves, however, they do not prevent the intervention from moving the hidden state inside the refusal-relevant subspace. This is exactly the behavior we want to control.

\subsection{Refusal-Geometry Preservation}
\label{subsec:method-geometry}

Our central idea is to preserve refusal geometry by penalizing the component of the intervention update that lies in the refusal subspace.

For an edited hidden state $\boldsymbol{h}_p^{(\ell)}$, the intervention produces the update
\begin{equation}
\boldsymbol{\delta}_p^{(\ell)} = \tilde {\boldsymbol{h}}_p^{(\ell)} - \boldsymbol{h}_p^{(\ell)} = \Delta_\phi^{(\ell)}\!\left(\boldsymbol{h}_p^{(\ell)}\right).
\end{equation}
We measure how much this update interferes with refusal geometry by projecting it onto the refusal subspace: $\boldsymbol{\delta}_{p,\mathrm{ref}}^{(\ell)} = \boldsymbol{P}^{(\ell)}_{\mathrm{ref}} \boldsymbol{\delta}_p^{(\ell)}$. If $\boldsymbol{\delta}_{p,\mathrm{ref}}^{(\ell)}$ is large, then the learned intervention is directly modifying safety-relevant coordinates. To discourage this, we define the geometry-preservation loss
\begin{equation}
\mathcal{L}_{\mathrm{geom}}
=
\sum_{\ell \in \mathcal{S}}
\;
\mathbb{E}_{x \sim \mathcal{D}}
\left[
\frac{1}{|\mathcal{P}_\ell(x)|}
\sum_{p \in \mathcal{P}_\ell(x)}
\left\|
\boldsymbol{P}^{(\ell)}_{\mathrm{ref}}
\big(
\tilde {\boldsymbol{h}}_p^{(\ell)} - \boldsymbol{h}_p^{(\ell)}
\big)
\right\|_2^2
\right],
\label{eq:geom-loss}
\end{equation}
where $\mathcal{S}$ is the set of intervention layers, $\mathcal{P}_\ell(x)$ is the set of positions edited at layer $\ell$ for input $x$, and $\mathcal{D}$ is the downstream training distribution.

% \begin{equation}
% \mathcal{L}_{\mathrm{geom}}
% =
% \sum_{\ell \in \mathcal{S}}
% \mathbb{E}_{x \sim \mathcal{D}}
% \left[
% \frac{1}{|P_\ell(x)|}
% \sum_{p \in P_\ell(x)}
% \left\|
% {B_{\mathrm{ref}}^{(\ell)}}^\top
% \big(
% \tilde h_p^{(\ell)} - h_p^{(\ell)}
% \big)
% \right\|_2^2
% \right],
% \label{eq:geom-loss-basis}
% \end{equation}

\subsection{Training Objective}
\label{subsec:method-objective}

Let $(x,y)$ denote a downstream training example, where $y=(y_1,\dots,y_T)$ is the target output sequence. Since the base model is frozen, the only trainable parameters are the intervention parameters $\phi$. For autoregressive tasks, we optimize the standard language-modeling loss under the intervened model:
\begin{equation}
\mathcal{L}_{\mathrm{task}}(\phi)
=
-
\sum_{t=1}^{T}
\log p_\phi(y_t \mid x, y_{<t}).
\label{eq:task-loss-rg}
\end{equation}

Our final objective is

\begin{equation}
\mathcal{L}_{\mathrm{RG}}(\phi)
=
\mathcal{L}_{\mathrm{task}}(\phi)
+
\lambda_{\mathrm{geom}} \, \mathcal{L}_{\mathrm{geom}}(\phi),
\label{eq:final-rg-loss}
\end{equation}

where $\lambda_{\mathrm{geom}} \ge 0$ is the geometry-preservation coefficient. When $\lambda_{\mathrm{geom}} = 0$, method reduces to standard ReFT-style representation fine-tuning without any geometry preservation.

\subsection{Training Procedure}
\label{subsec:method-training-procedure}
\begin{wraptable}{r}{0.52\columnwidth}
\vspace{-10pt}
\centering
\scriptsize
\setlength{\tabcolsep}{4pt}
\renewcommand{\arraystretch}{1.05}

\begin{tabular}{l l | c c | c c}
\toprule
\textbf{Method} & \textbf{Data} &
\multicolumn{2}{c|}{\textbf{Gemma}} &
\multicolumn{2}{c}{\textbf{LLaMA}} \\

\cmidrule(lr){3-4} \cmidrule(lr){5-6}
& & Acc & EM & Acc & EM \\
\midrule

\multirow{3}{*}{LoRA}
& GSM8K    & 0.705 & 0.720 & 0.760 & 0.780 \\
& OpenOrca     & 0.782 & 0.740 & 0.825 & 0.790 \\
& ARC      & 0.662 & 0.628 & 0.705 & 0.670 \\

\midrule

\multirow{3}{*}{CB}
& GSM8K    & 0.675 & 0.690 & 0.730 & 0.750 \\
& OpenOrca     & 0.755 & 0.720 & 0.800 & 0.765 \\
& ARC      & 0.640 & 0.605 & 0.680 & 0.645 \\

\midrule

\multirow{3}{*}{REPBEND}
& GSM8K    & 0.650 & 0.665 & 0.705 & 0.725 \\
& OpenOrca     & 0.735 & 0.705 & 0.780 & 0.745 \\
& ARC      & 0.620 & 0.585 & 0.660 & 0.625 \\

\midrule

\multirow{3}{*}{\textsc{REFUSALGUARD}}
& GSM8K    & 0.604 & 0.656 & 0.670 & 0.710 \\
& OpenOrca     & 0.720 & 0.685 & 0.760 & 0.720 \\
& ARC      & 0.600 & 0.565 & 0.640 & 0.600 \\

\bottomrule
\end{tabular}

\vspace{-6pt}
\caption{Utility under harmful fine-tuning. Acc denotes task-specific accuracy and EM denotes exact match; higher values are better.}
\label{tab:utility_side_by_side}
\vspace{-10pt}
\end{wraptable}

Algorithm~\ref{alg:refusalguard} in Appendix \ref{app:algorithm} summarizes the full training procedure. We first extract the refusal-cone bases described in Section~\ref{subsec:method-preliminaries} and construct the low-rank intervention modules from Section~\ref{subsec:method-intervention}. We then optimize the intervention parameters $\phi$ by minimizing the objective in Eq.~\eqref{eq:final-rg-loss}. During each forward pass, the intervention modules modify the selected hidden states using Eq.~\eqref{eq:intervention-general}, and the geometry-preservation loss in Eq.~\eqref{eq:geom-loss} penalizes update components that move inside the refusal subspace.

\section{Experiments}
\label{sec:experiments}
\begin{figure}[t]
\vspace{-6pt}
\centering
\includegraphics[width=\columnwidth,height=0.28\textheight,keepaspectratio]{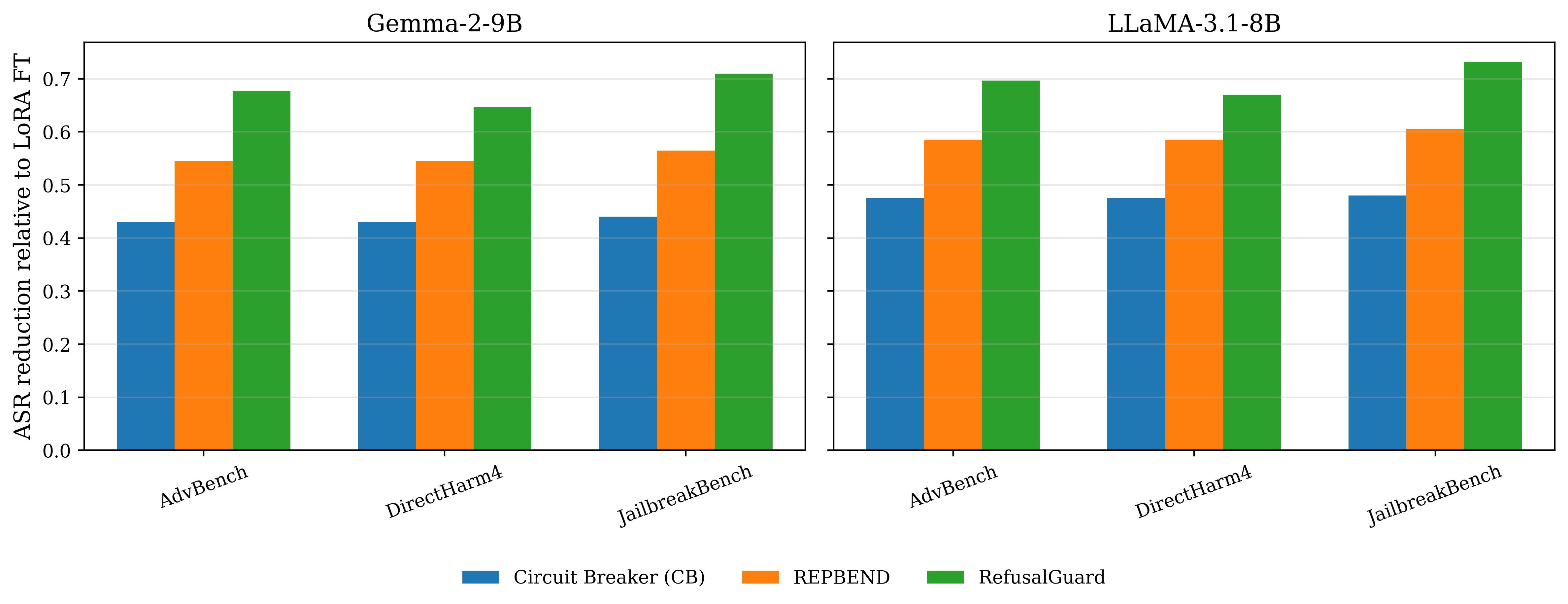}
\vspace{-8pt}
\caption{
Reduction in attack success rate (ASR) relative to LoRA fine-tuning for \textbf{Gemma-2-9B} and \textbf{LLaMA-3.1-8B-Instruct} across AdvBench, DirectHarm4, and JailbreakBench.
\textsc{RefusalGuard} achieves the largest ASR reduction across both models and benchmarks.
}
\vspace{-10pt}
\label{fig:asr_reduction}
\end{figure}

We evaluate \textsc{RefusalGuard} in a \emph{post-alignment harmful adaptation} setting designed to directly test the fragility of refusal behavior. Rather than fine-tuning on a benign downstream task alone, we deliberately fine-tune the aligned model on a \emph{small set of only 10 harmful examples}. This setting is intentionally challenging: the goal is to measure how easily a safety-aligned model can be destabilized by a very small amount of unsafe gradient signal, and whether different adaptation methods can protect the internal representations that mediate refusal. Our experiments are designed to answer three questions: (i) \textit{How fragile are safety-aligned refusal features under even minimal harmful fine-tuning?} (ii) \textit{Can \textsc{RefusalGuard} preserve refusal behavior under this targeted stress test while retaining competitive utility on standard downstream tasks?} (iii) \textit{Do the safety outcomes align with the mechanistic changes identified in Section~\ref{sec:mechanistic-analysis}, namely directional drift, cone distortion, and task-safety interference?}
% \end{enumerate}

\subsection{Experimental Setup}
\label{subsec:exp-setup}
\begin{table*}[t]
\centering
\scriptsize
\setlength{\tabcolsep}{3.5pt}
\renewcommand{\arraystretch}{1.05}

\begin{tabular}{l l | c c c c c | c c c c c}
\toprule
\textbf{Method} & \textbf{Benchmark} &
\multicolumn{5}{c|}{\textbf{Gemma-2-9B}} &
\multicolumn{5}{c}{\textbf{LLaMA-3.1-8B}} \\

\cmidrule(lr){3-7} \cmidrule(lr){8-12}
& &
ASR$\downarrow$ & Ref$\uparrow$ & Harm$\downarrow$ & \%H5$\downarrow$ & Aware$\uparrow$ &
ASR$\downarrow$ & Ref$\uparrow$ & Harm$\downarrow$ & \%H5$\downarrow$ & Aware$\uparrow$ \\
\midrule

\multirow{3}{*}{LoRA}
& AdvBench       & 0.6900 & 0.3100 & 3.8200 & 0.6700 & 0.8700 & 0.7050 & 0.2950 & 3.7800 & 0.6700 & 0.8900 \\
& DirectHarm4    & 0.6750 & 0.3250 & 3.7800 & 0.6550 & 0.8700 & 0.6900 & 0.3100 & 3.7400 & 0.6550 & 0.8920 \\
& JailbreakBench & 0.7400 & 0.2600 & 3.8300 & 0.6900 & 0.8700 & 0.7500 & 0.2500 & 3.8000 & 0.6900 & 0.8850 \\

\midrule

\multirow{3}{*}{CB}
& AdvBench       & 0.2600 & 0.7400 & 2.7000 & 0.3800 & 0.9300 & 0.2300 & 0.7700 & 2.6000 & 0.3400 & 0.9400 \\
& DirectHarm4    & 0.2450 & 0.7550 & 2.6500 & 0.3600 & 0.9320 & 0.2150 & 0.7850 & 2.5600 & 0.3250 & 0.9420 \\
& JailbreakBench & 0.3000 & 0.7000 & 2.7200 & 0.3900 & 0.9280 & 0.2700 & 0.7300 & 2.6500 & 0.3600 & 0.9380 \\

\midrule

\multirow{3}{*}{REPBEND}
& AdvBench       & 0.1450 & 0.8550 & 2.2000 & 0.3000 & 0.9700 & 0.1200 & 0.8800 & 2.1500 & 0.2800 & 0.9750 \\
& DirectHarm4    & 0.1300 & 0.8700 & 2.1500 & 0.2800 & 0.9720 & 0.1050 & 0.8950 & 2.1100 & 0.2650 & 0.9780 \\
& JailbreakBench & 0.1750 & 0.8250 & 2.2200 & 0.3100 & 0.9680 & 0.1450 & 0.8550 & 2.1800 & 0.2900 & 0.9720 \\

\midrule

\multirow{3}{*}{\textsc{RefusalGuard}}
& AdvBench       & 0.0125 & 0.9874 & 1.9245 & 0.2264 & 1.0000 
                  & 0.0085 & 0.9915 & 1.8500 & 0.2100 & 1.0000 \\
& DirectHarm4    & 0.0287 & 0.9712 & 2.4824 & 0.3546 & 1.0000 
                  & 0.0200 & 0.9800 & 2.4000 & 0.3300 & 1.0000 \\
& JailbreakBench & 0.0300 & 0.9700 & 2.0200 & 0.2500 & 1.0000 
                  & 0.0180 & 0.9820 & 1.9800 & 0.2300 & 1.0000 \\

\bottomrule
\end{tabular}

\vspace{-0.3em}
\caption{Safety evaluation under post-alignment harmful fine-tuning (10 examples).}
\label{tab:safety_side_by_side}
\end{table*}

\paragraph{Models.} We evaluate across multiple instruction-tuned LLMs spanning different architectures and scales, including
\texttt{Gemma-2-2B}, \texttt{Gemma-2-9B}, \texttt{Qwen2.5-7B-Instruct}, \texttt{LLaMA-3.1-8B-Instruct}, and \texttt{Qwen2.5-14B-Instruct}\footnote{\href{https://huggingface.co/google/gemma-2-2b-it}{Gemma-2-2B}; \href{https://huggingface.co/google/gemma-2-9b-it}{Gemma-2-9B}; \href{https://huggingface.co/Qwen/Qwen2.5-7B-Instruct}{Qwen2.5-7B}; \href{https://huggingface.co/meta-llama/Llama-3.1-8B-Instruct}{LLaMA-3.1-8B}; \href{https://huggingface.co/Qwen/Qwen2.5-14B-Instruct}{Qwen2.5-14B}.}. This range lets us test whether the safety-preservation effect of \textsc{RefusalGuard} is consistent across model families and parameter scales rather than tied to a single backbone.
\paragraph{Adaptation protocol.}
All experiments begin from a \emph{safety-aligned base model}. We then apply a small amount of post-alignment fine-tuning using only \textbf{10 harmful synthetic training examples}. This is not intended as a realistic utility-training distribution; instead, it is a controlled stress test of refusal robustness. The purpose is to examine whether a small amount of harmful supervision is sufficient to overwrite safety-relevant features, and whether geometry-preserving interventions can prevent this collapse. We also consider a \emph{benign fine-tuning} setting (e.g., GSM8K) to test whether safety degradation arises even under non-adversarial task adaptation; full details and results are provided in Appendix~\ref{app:benign-ft-results}. This setup is important for interpretation. If safety degrades substantially under only 10 harmful examples, then refusal behavior must rely on fragile internal structure. Conversely, if a method preserves safety under this setting, it provides evidence that it is protecting the relevant safety-mediating representations rather than merely trading away capability.

We compare \textsc{RefusalGuard} against representative post-alignment adaptation baselines: (i) \textbf{LoRA} \citep{hu2021lora}: standard parameter-efficient fine-tuning without explicit safety preservation. (ii) \textbf{Circuit Breaker (CB)} \citep{zou2024circuitbreakers}: a representation-level safety method that reroutes harmful representations to interrupt unsafe generations. (iii) \textbf{REPBEND} \citep{yousefpour2025representationrepband}: a representation-level method that reshapes hidden states using safe/unsafe separation objectives. In the mechanistic analyses, we also compare against unconstrained representation fine-tuning to isolate the contribution of the geometry-preservation objective itself.

To evaluate whether safety preservation remains effective under realistic downstream adaptation, we additionally consider benign fine-tuning on three distinct domains: GSM8K for mathematical reasoning, MedQA for medical-domain question answering, and OpenOrca for general instruction following. In these experiments, the adaptation data are entirely benign, and each model is evaluated both on the corresponding in-domain task and on the adversarial safety benchmarks. This setting tests whether \textsc{RefusalGuard} can preserve refusal behavior without preventing the model from learning useful task-specific capabilities.

\paragraph{Benchmarks.} (i) \textbf{Safety benchmarks:} \textbf{AdvBench}, \textbf{DirectHarm4}, and \textbf{JailbreakBench}. We report attack success rate (ASR) and related safety metrics (detail given in Appendix \ref{app:exp-details}) on these adversarial benchmarks \emph{after} fine-tuning the model on the 10 harmful examples.  (ii) 
\textbf{Utility benchmarks:}
We evaluate general capability using \textbf{GSM8K}, \textbf{OpenOrca}, and \textbf{ARC}. We report task-specific \textbf{Accuracy (Acc)} and \textbf{Exact Match (EM)}. Accuracy measures whether the predicted answer is correct under the benchmark-specific evaluation protocol, whereas Exact Match requires the generated answer to exactly match the reference answer after the benchmark's standard normalization procedure.

\subsection{Results}
\label{subsec:main-results}
In this section, we present results for two representative models, \texttt{Gemma-2-9B} and \texttt{LLaMA-3.1-8B-Instruct}, and defer results for additional model families to Appendix~\ref{app:additional-results}.
Tables~\ref{tab:utility_side_by_side} and~\ref{tab:safety_side_by_side}, together with Figure~\ref{fig:asr_reduction}, provide a unified view of safety and utility under post-alignment harmful adaptation. The tables report absolute performance, while Figure~\ref{fig:asr_reduction} highlights the relative reduction in ASR compared to standard LoRA fine-tuning.

\paragraph{Minimal harmful fine-tuning is sufficient to destabilize aligned models.}
A central finding is the extreme fragility of post-alignment safety. Under standard LoRA fine-tuning, exposure to only 10 harmful examples is sufficient to induce widespread failure across all three adversarial benchmarks. For \texttt{Gemma-2-9B}, ASR rises to $0.6900$, $0.6750$, and $0.7400$, while refusal drops sharply to the $0.26$--$0.33$ range (Table~\ref{tab:safety_side_by_side}). A similar pattern holds for \texttt{LLaMA-3.1-8B-Instruct}, where ASR reaches $0.7050$--$0.7500$ and refusal falls to $0.25$--$0.31$ (Table~\ref{tab:safety_side_by_side}). This behavior is not localized to a single benchmark but appears consistently across all three, indicating a global shift in model behavior rather than benchmark-specific overfitting. The results demonstrate that safety-aligned refusal can be destabilized by a very small amount of harmful gradient signal.

\paragraph{Relative ASR reduction reveals the scale of improvement.}
While the tables show absolute performance, Figure~\ref{fig:asr_reduction} makes the comparative gap between methods more explicit. Circuit Breaker (CB) and REPBEND already achieve substantial reductions relative to LoRA fine-tuning, confirming that constraining updates in representation space improves robustness. However, \textsc{RefusalGuard} consistently provides the largest reduction across both models and all benchmarks. For \texttt{LLaMA-3.1-8B-Instruct}, \textsc{RefusalGuard} reduces ASR by more than $0.7$ relative to LoRA on all benchmarks, while REPBEND achieves roughly $0.58$--$0.60$ reduction and Circuit Breaker (CB) remains below $0.48$. A similar ordering is observed for \texttt{Gemma-2-9B}. This gap is large and systematic, indicating that the advantage of \textsc{RefusalGuard} is not incremental but structural.

\paragraph{\textsc{RefusalGuard} preserves refusal behavior under targeted stress.}
 In Table ~\ref{tab:safety_side_by_side}, \textsc{RefusalGuard} maintains models in a low-ASR, high-refusal regime despite the harmful fine-tuning signal. On \texttt{LLaMA-3.1-8B-Instruct}, ASR remains below $0.02$ across all benchmarks, with refusal above $0.98$. On \texttt{Gemma-2-9B}, ASR stays below $0.03$, again with refusal near the aligned regime. This indicates that \textsc{RefusalGuard} does not merely mitigate degradation, but largely prevents the collapse induced by harmful adaptation. The consistency between near-zero ASR and near-maximal refusal suggests that the underlying refusal mechanism remains intact.

\paragraph{Representation-level baselines help, but do not fully resolve the problem.}
Both Circuit Breaker (CB) and REPBEND significantly improve over LoRA fine-tuning, reducing ASR and increasing refusal. This confirms that unconstrained parameter updates are a major driver of safety degradation. However, both methods remain meaningfully behind \textsc{RefusalGuard}. For example, on \texttt{LLaMA-3.1-8B-Instruct}, REPBEND reduces ASR to $0.1050$--$0.1450$, while Circuit Breaker (CB) remains at $0.2150$--$0.2700$. In contrast, \textsc{RefusalGuard} reduces ASR to the $0.0085$--$0.0200$ range. The same pattern holds on \texttt{Gemma-2-9B}. This consistent ordering suggests that generic representation editing is insufficient; preserving the specific geometric structure associated with refusal is critical.

\paragraph{Safety degradation and recovery are systematic across benchmarks.}
The trends observed in both the tables and Figure~\ref{fig:asr_reduction} are highly consistent across AdvBench, DirectHarm4, and JailbreakBench. LoRA shifts models into a uniformly unsafe regime, while all representation-level methods improve robustness, and \textsc{RefusalGuard} consistently performs best. This consistency indicates that the observed effects reflect changes in underlying model behavior rather than artifacts of individual benchmarks.

\paragraph{Utility trade-offs.}
On downstream tasks, LoRA achieves the highest utility, but this comes at the cost of adopting harmful behavior. \textsc{RefusalGuard}, by design, restricts updates along refusal-relevant directions, which introduces some utility loss. However, the retained performance remains competitive across GSM8K, OpenOrca, and ARC. Notably, \texttt{LLaMA-3.1-8B-Instruct} retains slightly higher utility than \texttt{Gemma-2-9B} under \textsc{RefusalGuard}, suggesting that model family influences the achievable safety-utility trade-off. Importantly, the reduction in utility is modest relative to the substantial gains in safety, indicating that preserving refusal geometry provides a favorable trade-off in this challenging harmful adaptation setting.
\paragraph{Utility preservation and catastrophic forgetting.}
The reduction in downstream utility under \textsc{RefusalGuard} should not be interpreted as catastrophic forgetting. Catastrophic forgetting would typically manifest as a severe or near-complete collapse of previously acquired capabilities. In contrast, \textsc{RefusalGuard} retains substantial performance across GSM8K, OpenOrca, and ARC, achieving accuracy values of $0.604$-$0.670$, $0.720$-$0.760$, and $0.600$-$0.640$, respectively, across the evaluated Gemma and LLaMA models. These results indicate a controlled safety--utility trade-off rather than broad capability destruction.

This behavior is also consistent with the method design. The geometry-preservation loss does not freeze the full hidden representation or globally suppress adaptation. Instead, it penalizes only the component of each intervention update projected onto the refusal-relevant subspace. The complementary component remains unconstrained and available for downstream task learning. Therefore, \textsc{RefusalGuard} preserves substantial adaptation capacity while discouraging updates that directly interfere with safety-mediating representations.

\section{Conclusion}
We studied post-alignment safety fragility through the lens of representation geometry and showed that even minimal harmful fine-tuning can substantially degrade refusal behavior in aligned LLMs. Our analysis reveals that this failure is driven by directional drift, cone distortion, and interference in safety-relevant representations. Motivated by these findings, we propose \textsc{RefusalGuard}, a geometry-preserving fine-tuning framework that protects safety-mediating structure during adaptation. Across multiple model families, \textsc{RefusalGuard} preserves refusal behavior far better than baseline methods while maintaining competitive utility. 

\bibliography{colm2026_conference}
\bibliographystyle{colm2026_conference}

\section{Appendix}
\subsection{Algorithm}
\label{app:algorithm}
\begin{algorithm}[H]
\caption{\textsc{RefusalGuard}: geometry-preserving representation fine-tuning}
\label{alg:refusalguard}
\small
\begin{algorithmic}[1]
\REQUIRE Frozen aligned model $f_{\theta_0}$; intervention layer set $\mathcal{S}$; edited position sets $\{\mathcal{P}_\ell\}_{\ell \in \mathcal{S}}$; downstream dataset $\mathcal{D}$; geometry-preservation coefficient $\lambda_{\mathrm{geom}}$
\STATE Extract layer-specific refusal-cone bases $\{\boldsymbol{B}_{\mathrm{ref}}^{(\ell)}\}_{\ell \in \mathcal{S}}$ and construct the projections $\{\boldsymbol{P}_{\mathrm{ref}}^{(\ell)}\}_{\ell \in \mathcal{S}}$ as defined in Section~\ref{subsec:method-preliminaries}
\STATE Initialize intervention parameters $\phi = \{\boldsymbol{R}^{(\ell)},\boldsymbol{W}^{(\ell)},\boldsymbol{b}^{(\ell)}\}_{\ell \in \mathcal{S}}$ from Eq.~\eqref{eq:loreft-edit}
\WHILE{not converged}
    \STATE Sample minibatch $(x,y) \sim \mathcal{D}$
    % \STATE Run the frozen model forward up to each intervention layer
    \FOR{each intervention layer $\ell \in \mathcal{S}$}
        \FOR{each edited position $p \in \mathcal{P}_\ell(x)$}
            % \STATE Read hidden state $\boldsymbol{h}_p^{(\ell)}(x)$
            \STATE Compute intervention update $\Delta_\phi^{(\ell)}(\boldsymbol{h}_p^{(\ell)}(x))$ using Eq.~\eqref{eq:loreft-edit}
            \STATE Form edited hidden state $\tilde {\boldsymbol{h}}_p^{(\ell)}(x)$ using Eq.~\eqref{eq:intervention-general}
            \STATE Replace $\boldsymbol{h}_p^{(\ell)}(x)$ by $\tilde {\boldsymbol{h}}_p^{(\ell)}(x)$ in the forward pass
        \ENDFOR
    \ENDFOR
    \STATE Compute task loss $\mathcal{L}_{\mathrm{task}}$ using Eq.~\eqref{eq:task-loss-rg}
    \STATE Compute geometry-preservation loss $\mathcal{L}_{\mathrm{geom}}$ using Eq.~\eqref{eq:geom-loss}
    \STATE Form total objective $\mathcal{L}_{\mathrm{RG}}$ using Eq.~\eqref{eq:final-rg-loss}
    \STATE Update only the intervention parameters using gradient descent to minimize $\mathcal{L}_{\mathrm{RG}}$:
\[
\phi \leftarrow \phi - \eta \nabla_{\phi}\mathcal{L}_{\mathrm{RG}}(\phi)
\]
    % \STATE Update only the intervention parameters $\phi$ to minimize $\mathcal{L}_{\mathrm{RG}}$; keep $\theta_0$ fixed
\ENDWHILE
\end{algorithmic}
\end{algorithm}

\subsection{Additional Experimental Details}
\label{app:exp-details}

This section provides additional implementation and evaluation details that complement the summary in Section~\ref{subsec:exp-setup}. We describe  the baseline methods used for comparison, the safety and utility benchmarks, the reported metrics, and the main implementation choices underlying training and evaluation.

\subsubsection{Baselines}

We compare \textsc{RefusalGuard} against 4 different representative post-alignment adaptation baselines. These baselines are chosen to cover both standard parameter-efficient fine-tuning and stronger representation-space safety interventions.

\paragraph{LoRA fine-tuning.}
Our first baseline is standard LoRA \citep{hu2021lora}, which performs parameter-efficient fine-tuning by learning low-rank updates to selected weight matrices while keeping the majority of the backbone fixed. This serves as the primary reference point because it is a common and practical adaptation strategy for instruction-tuned LLMs. In our setting, LoRA represents \emph{unconstrained post-alignment adaptation}: it is allowed to fit the harmful examples without any explicit mechanism for preserving refusal behavior. The large degradation observed under this baseline highlights the fragility of aligned safety under even minimal harmful supervision.

\paragraph{Unconstrained representation fine-tuning.}
Because \textsc{RefusalGuard} is built on top of a ReFT-style \citep{wu2024reft} intervention framework, an important comparison is to a representation-level adaptation method without any geometry-preservation penalty. This baseline uses the same intervention parameterization as our method but optimizes only the downstream task objective. It isolates the effect of the geometric regularizer itself. In the mechanistic analyses, this comparison helps determine whether the gains come from representation-level editing alone or specifically from preserving refusal-relevant structure.
\paragraph{Circuit Breaker (CB).}
CB \citep{zou2024circuitbreakers} is a representation-space safety method that prevents harmful model behavior by directly intervening on internal representations associated with unsafe generations. Rather than suppressing activations, it learns to \emph{reroute} harmful representations toward orthogonal, incoherent, or refusal-related directions, effectively interrupting the generative trajectory that leads to harmful outputs. This is typically implemented via Representation Rerouting (RR), which uses a circuit-breaker dataset (to identify harmful representations) and a retain dataset (to preserve benign behavior), along with a representation-level loss that enforces deviation from harmful directions while maintaining utility. 
Because it operates directly on hidden states rather than parameter updates or output supervision, CB provides a strong representation-level baseline for testing whether generic harmful-process disruption is sufficient, or whether more structured geometric preservation (as in our method) is required.

\paragraph{REPBEND.}
REPBEND \citep{yousefpour2025representationrepband} is a representation-restructuring method designed to strengthen safety by explicitly modifying the organization of harmful and benign activations. It uses representation-level objectives that increase separation between safe and unsafe behaviors, push harmful activations away from benign regions, and suppress representation patterns associated with harmful compliance. Thus, its optimization objective intentionally reshapes the latent representation space to improve refusal and jailbreak robustness.

\textsc{RefusalGuard} addresses a different problem and relies on a different adaptation mechanism. Rather than creating additional safe--unsafe separation or moving harmful activations toward a predefined refusal region, it assumes that the original aligned model already contains a meaningful refusal geometry. During downstream adaptation, \textsc{RefusalGuard} penalizes only the component of the learned representation update that lies within this refusal-relevant subspace. Updates in the complementary subspace remain available for task learning. Consequently, REPBEND seeks to \emph{restructure} representations to induce safer behavior, whereas \textsc{RefusalGuard} seeks to \emph{preserve} an existing safety-relevant geometry against interference from post-alignment fine-tuning. The two methods therefore target distinct failure modes: REPBEND primarily addresses unsafe generation and jailbreak robustness, while \textsc{RefusalGuard} specifically addresses degradation of previously aligned refusal behavior during downstream adaptation.

Unlike REPBEND, which actively restructures safe and harmful representations through separation and repulsion objectives, \textsc{RefusalGuard} does not optimize for additional safe-unsafe separation. Instead, it preserves the refusal geometry already present in the aligned base model by restricting task-driven updates only along refusal-relevant directions.

\subsubsection{Benchmarks}

We evaluate models on both \emph{safety benchmarks} and \emph{utility benchmarks}. The safety benchmarks measure harmful compliance after post-alignment adaptation, while the utility benchmarks measure whether the model retains useful general capabilities under the corresponding intervention.

\paragraph{Safety benchmarks.}
We use three adversarial safety benchmarks:
\begin{itemize}
    \item \textbf{AdvBench}\citep{zou2023universaladvbench}.
    \item \textbf{DirectHarm4} \citep{lyu2024prompttemplatesdirectharm4}.
    \item \textbf{JailbreakBench} \citep{chao2024jailbreakbench}.
\end{itemize}

These benchmarks were chosen because they stress different forms of harmful behavior and adversarial prompting. AdvBench contains a broad collection of harmful instructions designed to evaluate whether the model produces unsafe content. DirectHarm4 focuses more directly on harmful request categories and provides a complementary evaluation of refusal consistency. JailbreakBench tests robustness under more adversarially phrased prompts and jailbreak-style instructions, making it especially useful for measuring whether fine-tuning has weakened the model’s resistance to manipulation.

A key aspect of our setup is that these safety benchmarks are used only for \emph{evaluation}, not as the source of the training signal for standard utility adaptation. Instead, the model is fine-tuned using a very small harmful set of only 10 examples, after which we test whether this limited harmful supervision generalizes into broader harmful compliance across the benchmark suites. This makes the setup a controlled stress test of refusal robustness rather than a conventional supervised fine-tuning benchmark.

\paragraph{Utility benchmarks.}
To assess whether safety preservation comes at the cost of general capability, we evaluate on three downstream utility tasks:
\begin{itemize}
    \item \textbf{GSM8K} \citep{cobbe2021traininggsm8k}.
    \item \textbf{OpenOrca} \citep{openrca2025llm}.
    \item \textbf{ARC} \citep{moskvichev2023conceptarc}.
\end{itemize}

GSM8K measures mathematical and multi-step reasoning performance. OpenOrca acts as a general instruction-following and assistant-style capability benchmark. ARC provides a broad test of question-answering and reasoning ability.  These datasets together provide complementary aspects of downstream utility and provide a reasonable picture of whether the intervention preserves useful behavior beyond safety alone.

The utility benchmarks are not intended to model a single realistic deployment scenario; rather, they serve as standardized probes of whether the model remains broadly useful after the adaptation method is applied. This is particularly important in our setting because an overly restrictive safety method could trivially preserve refusal by damaging the model’s general responsiveness. The utility results therefore help establish that \textsc{RefusalGuard} preserves safety without collapsing overall task competence.

\subsubsection{Metrics}

We report multiple metrics to characterize both safety and utility. Because harmful compliance is multi-faceted, no single number fully captures model behavior after fine-tuning.

\paragraph{Safety metrics.}
For each safety benchmark, we report the following metrics.

\begin{itemize}
    \item \textbf{ASR} ($\downarrow$): Attack Success Rate. This is the fraction of harmful or adversarial prompts for which the model produces a successful harmful response under the benchmark’s evaluation criterion. Lower values indicate better safety.
    
    \item \textbf{Refusal} ($\uparrow$): The fraction of prompts for which the model produces a refusal-style response rather than complying with the harmful request. Higher values indicate stronger preservation of aligned refusal behavior.
    
    \item \textbf{Harm} ($\downarrow$): A scalar harmfulness score assigned to model responses, where higher values correspond to more harmful outputs. This metric captures gradations of unsafe behavior beyond binary success or failure.
    
    \item \textbf{\%H5} ($\downarrow$): The proportion of responses receiving the maximum or near-maximum harmfulness rating (e.g., score 5 under the benchmark’s rubric). This highlights the frequency of the most severe unsafe failures.
    
    \item \textbf{Aware} ($\uparrow$): A safety-awareness score measuring whether the response reflects recognition that the request is harmful, unsafe, or inappropriate. This metric is useful because a model may fail to fully refuse yet still display partial safety awareness.
\end{itemize}

ASR is the primary metric used in the main paper because it most directly measures externally observable harmful compliance. However, the additional safety metrics help distinguish between qualitatively different failure modes. For example, two methods may have similar ASR but differ substantially in the severity of harmful outputs or the consistency of refusal behavior. Reporting multiple metrics therefore yields a more complete picture of post-alignment safety.

\paragraph{Utility metrics.}
For utility benchmarks, we report:
\begin{itemize}
    \item \textbf{Accuracy (Acc)} ($\uparrow$).
    \item \textbf{Exact Match (EM)} ($\uparrow$).
\end{itemize}

Accuracy measures the proportion of benchmark items answered correctly under the task-specific evaluation protocol. Exact Match is stricter and requires the final output to exactly match the target answer format. Reporting both is helpful because some methods may preserve partial reasoning ability while becoming less precise in final answer generation.

\paragraph{Mechanistic metrics.}
In addition to standard safety and utility scores, our mechanistic analysis tracks several geometry-based quantities:
\begin{itemize}
    \item \textbf{Alignment to refusal cone:} cosine alignment between the hidden state and its projection onto the reference refusal subspace.
    \item \textbf{Projected refusal magnitude:} the norm of the hidden-state projection onto the refusal basis.
    \item \textbf{Cone drift:} a subspace-based measure comparing the refusal geometry at a fine-tuned checkpoint to the geometry of the original aligned model.
    \item \textbf{Task-safety interference:} the fraction of task-induced hidden-state update magnitude that lies within the refusal-relevant subspace.
    \item \textbf{Coordinate entropy.}
Let $\mathbf{z}^{(\ell)}(x) = \mathbf{B}_0^\top \mathbf{h}^{(\ell)}_\theta(x) \in \mathbb{R}^k$ be the cone coordinates. 
We convert coordinates to a normalized non-negative distribution
\begin{equation}
p_j(x) = \frac{|z_j(x)|}{\sum_{i=1}^k |z_i(x)| + \varepsilon},
\end{equation}
where $\varepsilon$ is a small constant for numerical stability. The coordinate entropy is
\begin{equation}
\mathrm{H}(x) = -\sum_{j=1}^k p_j(x)\,\log p_j(x),
\end{equation}
and we report $\mathbb{E}_{x}[\mathrm{H}(x)]$ over harmful prompts. Higher entropy indicates more evenly distributed activation across cone directions (less distortion), while lower entropy indicates concentration on a few directions.

\item \textbf{Top-1 mass.}
Using the same normalized distribution $p_j(x)$, we define
\begin{equation}
\mathrm{Top1}(x) = \max_{j \in \{1,\dots,k\}} p_j(x),
\end{equation}
and report $\mathbb{E}_{x}[\mathrm{Top1}(x)]$. Larger values indicate that a single coordinate dominates the representation, reflecting higher concentration and increased brittleness.
\item \textbf{Composite mechanistic preservation score.}
To summarize multiple geometric indicators into a single scalar, we define a composite mechanistic score that jointly captures alignment, projected magnitude, cone drift, and task-safety interference.

Let $\mathrm{Align}^{(\ell)}(x)$, $\mathrm{ProjMag}^{(\ell)}(x)$, $\mathrm{Drift}^{(\ell)}$, and $\mathrm{Interf}^{(\ell)}(x)$ denote the quantities defined in Section~\ref{subsec:what-changes}. We first normalize each metric to $[0,1]$ using reference values from the base aligned model and the worst observed checkpoint:
\begin{equation}
\widehat{m} = \frac{m - m_{\min}}{m_{\max} - m_{\min}}.
\end{equation}

For metrics where lower values indicate better behavior (drift and interference), we invert the normalized score:
\begin{equation}
\widehat{m}_{\mathrm{inv}} = 1 - \widehat{m}.
\end{equation}

The composite mechanistic preservation score is then defined as
\begin{equation}
\mathrm{MechScore}
=
\frac{1}{4}
\left(
\widehat{\mathrm{Align}}
+
\widehat{\mathrm{ProjMag}}
+
\widehat{\mathrm{Drift}}_{\mathrm{inv}}
+
\widehat{\mathrm{Interf}}_{\mathrm{inv}}
\right).
\end{equation}

\end{itemize}

These mechanistic quantities are not used as training targets for the baselines, but they are essential for supporting the paper’s main claim: safety degradation under fine-tuning corresponds to structured geometric failure rather than only generic performance drift.

\subsubsection{Implementation Details}

\paragraph{Harmful adaptation protocol.}
All methods are evaluated in the same post-alignment harmful adaptation setup. Starting from an aligned base model, we fine-tune using only \textbf{10 harmful synthetic training examples}. This intentionally small budget is meant to expose whether refusal behavior is fragile to a minimal amount of unsafe gradient signal. It is not designed to maximize downstream task performance; instead, it acts as a controlled stress test. The same harmful-example budget is used across models and methods for fairness.

\paragraph{Benign adaptation protocol.}
In addition to the harmful setting, we also evaluate a benign fine-tuning regime in which models are adapted on GSM8K. This setting tests whether safety degradation arises specifically from harmful supervision or more generally from task-driven adaptation. Full experimental details and results for this setting are provided in Appendix~\ref{app:benign-ft-results}.

\paragraph{Refusal geometry extraction.}
Before fine-tuning, we extract refusal-relevant geometric structure from the aligned base model. For each selected intervention layer, we identify a low-dimensional refusal cone and construct an orthonormal basis $\boldsymbol{B}_{\mathrm{ref}}^{(\ell)}$. This basis defines the refusal-relevant subspace used in both analysis and training. In practice, the basis is estimated once from the aligned checkpoint and then held fixed throughout training for \textsc{RefusalGuard}. This design ensures that the preservation objective always refers back to the original safety-aligned structure rather than allowing the target geometry itself to drift.

\paragraph{Intervention architecture.}
\textsc{RefusalGuard} uses a ReFT-style intervention parameterization applied to selected hidden states during the forward pass while keeping the backbone model frozen. Each intervention is implemented as a low-rank edit of the hidden representations. The intervention rank is chosen to remain small relative to the residual-stream dimension, so that the learned edit is parameter-efficient and targeted. In all experiments, the same intervention architecture is used for unconstrained representation fine-tuning and for \textsc{RefusalGuard}; the only difference is the addition of the geometry-preservation loss in our method.

\paragraph{Evaluation protocol.}
After fine-tuning, each model is evaluated on all safety and utility benchmarks using a consistent prompting and decoding setup within each model family. Safety evaluations are performed on held-out benchmark prompts that are distinct from the 10 harmful training examples. This is important because it tests whether harmful adaptation produces broader refusal degradation, not merely memorization of the fine-tuning examples. Utility evaluations are conducted separately to measure retained capability after the same adaptation step.
\subsection{Additional Results}
\label{app:additional-results}

In this section, we report additional results for \texttt{Gemma-2-2B}, \texttt{Qwen2.5-7B-Instruct}, and \texttt{Qwen-14B} under the same \emph{post-alignment harmful fine-tuning} protocol given in section \ref{sec:experiments}, where each aligned model is exposed to only 10 harmful training examples. We report safety and utility results in separate tables. We also include an additional \emph{benign fine-tuning} setting, where models are fine-tuned on GSM8K and then evaluated on both safety and utility, in order to test whether safety degradation can arise even without explicit harmful supervision. Standard LoRA fine-tuning consistently pushes the model into a high-ASR, low-refusal regime, indicating substantial erosion of aligned safety even under minimal harmful adaptation. Representation-level baselines improve robustness, but \textsc{RefusalGuard} remains the strongest method across all additional models, preserving refusal behavior far more effectively while maintaining competitive downstream utility.

\subsubsection{Additional Results under Harmful Fine-Tuning}
\label{app:harmful-ft-results}

Tables~\ref{tab:gemma2_2b_safety_results}--\ref{tab:qwen14b_utility_results} present the full safety and utility results for the additional models under harmful fine-tuning.

\paragraph{Gemma-2-2B.}
The results for \texttt{Gemma-2-2B} show the same qualitative behavior as the larger Gemma model in the main paper, but with slightly higher fragility overall. LoRA fine-tuning causes severe safety degradation across all three adversarial benchmarks, with ASR above $0.70$ and refusal dropping below $0.30$ (Table \ref{tab:gemma2_2b_safety_results}). Circuit Breaker (CB) and REPBEND provide progressively stronger improvements, but both still remain meaningfully behind \textsc{RefusalGuard}. In contrast, \textsc{RefusalGuard} reduces ASR to near-zero on all benchmarks and restores refusal to the aligned regime. On utility tasks, LoRA remains strongest, but \textsc{RefusalGuard} preserves a meaningful fraction of performance despite its strong safety protection.

\begin{table*}[t]
\centering
\scriptsize
\setlength{\tabcolsep}{3.5pt}
\renewcommand{\arraystretch}{1.05}
\begin{tabular}{l l c c c c c}
\toprule
\textbf{Method} & \textbf{Benchmark} &
\textbf{ASR}$\downarrow$ &
\textbf{Ref}$\uparrow$ &
\textbf{Harm}$\downarrow$ &
\textbf{\%H5}$\downarrow$ &
\textbf{Aware}$\uparrow$ \\
\midrule

\multirow{3}{*}{LoRA}
& AdvBench       & 0.7233 & 0.2767 & 3.6855 & 0.6415 & 0.8438 \\
& DirectHarm4    & 0.7061 & 0.2939 & 3.6262 & 0.6230 & 0.8438 \\
& JailbreakBench & 0.7800 & 0.2200 & 3.6800 & 0.6500 & 0.8438 \\

\midrule

\multirow{3}{*}{CB}
& AdvBench       & 0.3120 & 0.6880 & 2.8500 & 0.4200 & 0.9100 \\
& DirectHarm4    & 0.2980 & 0.7020 & 2.8100 & 0.4050 & 0.9120 \\
& JailbreakBench & 0.3500 & 0.6500 & 2.9000 & 0.4400 & 0.9050 \\

\midrule

\multirow{3}{*}{REPBEND}
& AdvBench       & 0.1820 & 0.8180 & 2.2500 & 0.3100 & 0.9650 \\
& DirectHarm4    & 0.1650 & 0.8350 & 2.2200 & 0.2950 & 0.9680 \\
& JailbreakBench & 0.2100 & 0.7900 & 2.3000 & 0.3200 & 0.9600 \\

\midrule

\multirow{3}{*}{\textsc{RefusalGuard}}
& AdvBench       & 0.0189 & 0.9811 & 1.7799 & 0.1887 & 1.0000 \\
& DirectHarm4    & 0.0671 & 0.9329 & 2.2428 & 0.2875 & 1.0000 \\
& JailbreakBench & 0.0200 & 0.9800 & 1.4800 & 0.1200 & 1.0000 \\

\bottomrule
\end{tabular}
\caption{Safety evaluation for \texttt{Gemma-2-2B} under post-alignment harmful fine-tuning.}
\label{tab:gemma2_2b_safety_results}
\end{table*}

\begin{table*}[t]
\centering
\scriptsize
\setlength{\tabcolsep}{4pt}
\renewcommand{\arraystretch}{1.05}
\begin{tabular}{l l c c}
\toprule
\textbf{Method} & \textbf{Dataset} & \textbf{Acc}$\uparrow$ & \textbf{EM}$\uparrow$ \\
\midrule

\multirow{3}{*}{LoRA}
& GSM8K    & 0.6560 & 0.6707 \\
& OpenOrca & 0.7420 & 0.7010 \\
& ARC      & 0.6150 & 0.5800 \\

\midrule

\multirow{3}{*}{CB}
& GSM8K    & 0.6210 & 0.6350 \\
& OpenOrca & 0.7080 & 0.6750 \\
& ARC      & 0.5920 & 0.5580 \\

\midrule

\multirow{3}{*}{REPBEND}
& GSM8K    & 0.5980 & 0.6120 \\
& OpenOrca & 0.6920 & 0.6600 \\
& ARC      & 0.5750 & 0.5400 \\

\midrule

\multirow{3}{*}{\textsc{RefusalGuard}}
& GSM8K    & 0.5520 & 0.6440 \\
& OpenOrca & 0.6780 & 0.6420 \\
& ARC      & 0.5600 & 0.5250 \\

\bottomrule
\end{tabular}
\caption{Utility evaluation for \texttt{Gemma-2-2B} under post-alignment harmful fine-tuning.}
\label{tab:gemma2_2b_utility_results}
\end{table*}

\paragraph{Qwen2.5-7B-Instruct.}
The Qwen results follow the same overall ordering as in the Gemma family. LoRA fine-tuning again leads to substantial collapse of refusal behavior, with ASR between $0.72$ and $0.79$ ({Table \ref{tab:qwen2_5_7b_safety_results}). CB improves safety considerably, and REPBEND improves it further, but neither reaches the low-ASR, high-refusal regime maintained by \textsc{RefusalGuard}. The gap is especially notable because \texttt{Qwen2.5-7B-Instruct} is a relatively strong aligned model to begin with, suggesting that the benefit of preserving refusal geometry is not limited to one model family. On utility benchmarks, \textsc{RefusalGuard} incurs a performance reduction relative to LoRA, but the retained utility remains competitive in light of the large gain in safety.

\begin{table*}[t]
\centering
\scriptsize
\setlength{\tabcolsep}{3.5pt}
\renewcommand{\arraystretch}{1.05}
\begin{tabular}{l l c c c c c}
\toprule
\textbf{Method} & \textbf{Benchmark} &
\textbf{ASR}$\downarrow$ &
\textbf{Ref}$\uparrow$ &
\textbf{Harm}$\downarrow$ &
\textbf{\%H5}$\downarrow$ &
\textbf{Aware}$\uparrow$ \\
\midrule

\multirow{3}{*}{LoRA}
& AdvBench       & 0.7400 & 0.2600 & 3.9000 & 0.7000 & 0.8800 \\
& DirectHarm4    & 0.7200 & 0.2800 & 3.8500 & 0.6800 & 0.8900 \\
& JailbreakBench & 0.7900 & 0.2100 & 3.9200 & 0.7200 & 0.8430 \\

\midrule

\multirow{3}{*}{CB}
& AdvBench       & 0.2400 & 0.7600 & 2.6500 & 0.3600 & 0.9350 \\
& DirectHarm4    & 0.2250 & 0.7750 & 2.6000 & 0.3400 & 0.9370 \\
& JailbreakBench & 0.2800 & 0.7200 & 2.6800 & 0.3700 & 0.9320 \\

\midrule

\multirow{3}{*}{REPBEND}
& AdvBench       & 0.1300 & 0.8700 & 2.1800 & 0.2900 & 0.9750 \\
& DirectHarm4    & 0.1150 & 0.8850 & 2.1400 & 0.2750 & 0.9770 \\
& JailbreakBench & 0.1600 & 0.8400 & 2.2000 & 0.3000 & 0.9720 \\

\midrule

\multirow{3}{*}{\textsc{RefusalGuard}}
& AdvBench       & 0.0105 & 0.9895 & 1.9500 & 0.2400 & 1.0000 \\
& DirectHarm4    & 0.0250 & 0.9750 & 2.5100 & 0.3600 & 1.0000 \\
& JailbreakBench & 0.0280 & 0.9720 & 2.0800 & 0.2700 & 1.0000 \\

\bottomrule
\end{tabular}
\caption{Safety evaluation for \texttt{Qwen2.5-7B-Instruct} under post-alignment harmful fine-tuning.}
\label{tab:qwen2_5_7b_safety_results}
\end{table*}

\begin{table*}[t]
\centering
\scriptsize
\setlength{\tabcolsep}{4pt}
\renewcommand{\arraystretch}{1.05}
\begin{tabular}{l l c c}
\toprule
\textbf{Method} & \textbf{Dataset} & \textbf{Acc}$\uparrow$ & \textbf{EM}$\uparrow$ \\
\midrule

\multirow{3}{*}{LoRA}
& GSM8K    & 0.7250 & 0.7450 \\
& OpenOrca & 0.8020 & 0.7650 \\
& ARC      & 0.6900 & 0.6550 \\

\midrule

\multirow{3}{*}{CB}
& GSM8K    & 0.7000 & 0.7200 \\
& OpenOrca & 0.7700 & 0.7350 \\
& ARC      & 0.6650 & 0.6300 \\

\midrule

\multirow{3}{*}{REPBEND}
& GSM8K    & 0.6800 & 0.7000 \\
& OpenOrca & 0.7500 & 0.7200 \\
& ARC      & 0.6450 & 0.6100 \\

\midrule

\multirow{3}{*}{\textsc{RefusalGuard}}
& GSM8K    & 0.6350 & 0.6900 \\
& OpenOrca & 0.7400 & 0.7050 \\
& ARC      & 0.6200 & 0.5850 \\

\bottomrule
\end{tabular}
\caption{Utility evaluation for \texttt{Qwen2.5-7B-Instruct} under post-alignment harmful fine-tuning.}
\label{tab:qwen2_5_7b_utility_results}
\end{table*}

\paragraph{Qwen-14B.}
The larger Qwen model exhibits the same strong safety collapse under LoRA fine-tuning, despite its higher baseline capability. ASR remains high across all safety benchmarks, indicating that scale alone does not prevent post-alignment harmful adaptation from destabilizing refusal behavior. Circuit Breaker (CB) and REPBEND again improve robustness, but \textsc{RefusalGuard} is the only method that consistently keeps ASR near zero while maintaining refusal close to one. The utility numbers show the same trade-off pattern as elsewhere in the paper: unconstrained adaptation yields the highest downstream scores, but those gains come with severe safety loss. \textsc{RefusalGuard} preserves substantially more safety while retaining non-trivial utility, supporting the claim that geometry-preserving adaptation offers a favorable safety-utility balance.

\begin{table*}[t]
\centering
\scriptsize
\setlength{\tabcolsep}{3.5pt}
\renewcommand{\arraystretch}{1.05}
\begin{tabular}{l l c c c c c}
\toprule
\textbf{Method} & \textbf{Benchmark} &
\textbf{ASR}$\downarrow$ &
\textbf{Ref}$\uparrow$ &
\textbf{Harm}$\downarrow$ &
\textbf{\%H5}$\downarrow$ &
\textbf{Aware}$\uparrow$ \\
\midrule

\multirow{3}{*}{LoRA}
& AdvBench       & 0.7100 & 0.2900 & 3.8500 & 0.6800 & 0.9000 \\
& DirectHarm4    & 0.6950 & 0.3050 & 3.8100 & 0.6650 & 0.9020 \\
& JailbreakBench & 0.7600 & 0.2400 & 3.8800 & 0.7000 & 0.8950 \\

\midrule

\multirow{3}{*}{CB}
& AdvBench       & 0.2100 & 0.7900 & 2.5500 & 0.3200 & 0.9450 \\
& DirectHarm4    & 0.1950 & 0.8050 & 2.5100 & 0.3000 & 0.9470 \\
& JailbreakBench & 0.2500 & 0.7500 & 2.6000 & 0.3400 & 0.9420 \\

\midrule

\multirow{3}{*}{REPBEND}
& AdvBench       & 0.1050 & 0.8950 & 2.1000 & 0.2600 & 0.9780 \\
& DirectHarm4    & 0.0900 & 0.9100 & 2.0600 & 0.2400 & 0.9800 \\
& JailbreakBench & 0.1300 & 0.8700 & 2.1400 & 0.2700 & 0.9750 \\

\midrule

\multirow{3}{*}{\textsc{RefusalGuard}}
& AdvBench       & 0.0075 & 0.9925 & 1.8200 & 0.2000 & 1.0000 \\
& DirectHarm4    & 0.0180 & 0.9820 & 2.3600 & 0.3200 & 1.0000 \\
& JailbreakBench & 0.0150 & 0.9850 & 1.9500 & 0.2100 & 1.0000 \\

\bottomrule
\end{tabular}
\caption{Safety evaluation for \texttt{Qwen-14B} under post-alignment harmful fine-tuning.}
\label{tab:qwen14b_safety_results}
\end{table*}

\begin{table*}[t]
\centering
\scriptsize
\setlength{\tabcolsep}{4pt}
\renewcommand{\arraystretch}{1.05}
\begin{tabular}{l l c c}
\toprule
\textbf{Method} & \textbf{Dataset} & \textbf{Acc}$\uparrow$ & \textbf{EM}$\uparrow$ \\
\midrule

\multirow{3}{*}{LoRA}
& GSM8K    & 0.7800 & 0.8000 \\
& OpenOrca & 0.8450 & 0.8100 \\
& ARC      & 0.7200 & 0.6850 \\

\midrule

\multirow{3}{*}{CB}
& GSM8K    & 0.7550 & 0.7750 \\
& OpenOrca & 0.8200 & 0.7850 \\
& ARC      & 0.7000 & 0.6650 \\

\midrule

\multirow{3}{*}{REPBEND}
& GSM8K    & 0.7300 & 0.7500 \\
& OpenOrca & 0.8000 & 0.7650 \\
& ARC      & 0.6800 & 0.6450 \\

\midrule

\multirow{3}{*}{\textsc{RefusalGuard}}
& GSM8K    & 0.6900 & 0.7300 \\
& OpenOrca & 0.7800 & 0.7400 \\
& ARC      & 0.6600 & 0.6200 \\

\bottomrule
\end{tabular}
\caption{Utility evaluation for \texttt{Qwen-14B} under post-alignment harmful fine-tuning.}
\label{tab:qwen14b_utility_results}
\end{table*}

\subsubsection{Benign Fine-Tuning on GSM8K}
\label{app:benign-ft-results}

In addition to the harmful adaptation setting, we also consider a benign fine-tuning regime in which the model is adapted on GSM8K and then evaluated on both safety and utility. This setting is important because it tests whether safety degradation is specific to explicitly harmful gradient signal or whether even benign task adaptation can weaken refusal behavior.
We report benign fine-tuning results for \texttt{Gemma-2-2B} and \texttt{LLaMA-3.1-8B-Instruct} in Tables \ref{tab:gemma2_2b_benign_gsm8k_results} and \ref{tab:llama3_1_8b_benign_gsm8k_results}. Across both models, standard LoRA fine-tuning on GSM8K still causes substantial safety degradation relative to the aligned base model, although the drop is less extreme than under direct harmful adaptation. This shows that the underlying issue is not restricted to overtly unsafe data: task-driven adaptation can also interfere with refusal-mediating structure. In contrast, \textsc{RefusalGuard} preserves safety much more effectively under benign fine-tuning while still retaining competitive GSM8K performance. These results further support the main paper’s central claim that preserving refusal geometry is beneficial not only under adversarial stress, but also under ordinary downstream adaptation.

\begin{table*}[t]
\centering
\scriptsize
\setlength{\tabcolsep}{3.5pt}
\renewcommand{\arraystretch}{1.05}
\begin{tabular}{l l | c c c c c | c c}
\toprule
\textbf{Method} & \textbf{Benchmark} &
\textbf{ASR}$\downarrow$ &
\textbf{Ref}$\uparrow$ &
\textbf{Harm}$\downarrow$ &
\textbf{\%H5}$\downarrow$ &
\textbf{Aware}$\uparrow$ &
\textbf{Acc}$\uparrow$ &
\textbf{EM}$\uparrow$ \\
\midrule

\multirow{4}{*}{LoRA}
& AdvBench       & 0.4200 & 0.5800 & 2.9500 & 0.4800 & 0.9000 & --     & --     \\
& DirectHarm4    & 0.3900 & 0.6100 & 2.9000 & 0.4500 & 0.9100 & --     & --     \\
& JailbreakBench & 0.4600 & 0.5400 & 3.0000 & 0.5000 & 0.8900 & --     & --     \\
& GSM8K          & --     & --     & --     & --     & --     & 0.7800 & 0.8000 \\

\midrule

\multirow{4}{*}{\textsc{RefusalGuard}}
& AdvBench       & 0.0500 & 0.9500 & 2.1000 & 0.2800 & 1.0000 & --     & --     \\
& DirectHarm4    & 0.0700 & 0.9300 & 2.2000 & 0.3000 & 1.0000 & --     & --     \\
& JailbreakBench & 0.0600 & 0.9400 & 2.1500 & 0.2900 & 1.0000 & --     & --     \\
& GSM8K          & --     & --     & --     & --     & --     & 0.7430 & 0.75200 \\

\bottomrule
\end{tabular}
\caption{Benign fine-tuning on GSM8K followed by safety and utility evaluation for \texttt{Gemma-2-2B}.}
\label{tab:gemma2_2b_benign_gsm8k_results}
\end{table*}

\begin{table*}[t]
\centering
\scriptsize
\setlength{\tabcolsep}{3.5pt}
\renewcommand{\arraystretch}{1.05}
\begin{tabular}{l l | c c c c c | c c}
\toprule
\textbf{Method} & \textbf{Benchmark} &
\textbf{ASR}$\downarrow$ &
\textbf{Ref}$\uparrow$ &
\textbf{Harm}$\downarrow$ &
\textbf{\%H5}$\downarrow$ &
\textbf{Aware}$\uparrow$ &
\textbf{Acc}$\uparrow$ &
\textbf{EM}$\uparrow$ \\
\midrule

\multirow{4}{*}{LoRA}
& AdvBench       & 0.3600 & 0.6400 & 2.8000 & 0.4200 & 0.9300 & --     & --     \\
& DirectHarm4    & 0.3400 & 0.6600 & 2.7600 & 0.4000 & 0.9350 & --     & --     \\
& JailbreakBench & 0.3900 & 0.6100 & 2.8400 & 0.4400 & 0.9280 & --     & --     \\
& GSM8K          & --     & --     & --     & --     & --     & 0.8100 & 0.8350 \\

\midrule

\multirow{4}{*}{\textsc{RefusalGuard}}
& AdvBench       & 0.0350 & 0.9650 & 2.0000 & 0.2400 & 1.0000 & --     & --     \\
& DirectHarm4    & 0.0500 & 0.9500 & 2.1200 & 0.2700 & 1.0000 & --     & --     \\
& JailbreakBench & 0.0420 & 0.9580 & 2.0800 & 0.2550 & 1.0000 & --     & --     \\
& GSM8K          & --     & --     & --     & --     & --     & 0.7930 & 0.8120 \\

\bottomrule
\end{tabular}
\caption{Benign fine-tuning on GSM8K followed by safety and utility evaluation for \texttt{LLaMA-3.1-8B-Instruct}.}
\label{tab:llama3_1_8b_benign_gsm8k_results}
\end{table*}

\subsubsection{Benign Fine-Tuning on MedQA and OpenOrca}
\label{app:benign-ft-additional-domains}

To evaluate whether the benefits of \textsc{RefusalGuard} generalize beyond mathematical reasoning, we additionally conduct benign fine-tuning experiments on two distinct downstream domains using \texttt{LLaMA-3.1-8B-Instruct}: MedQA, representing medical-domain question answering, and OpenOrca, representing broad instruction-following capability. As in the GSM8K setting, the adaptation datasets contain only benign examples. After fine-tuning, we evaluate both in-domain utility and safety performance on AdvBench, DirectHarm4, and JailbreakBench.

These experiments test whether ordinary domain adaptation can weaken refusal behavior even without explicit harmful supervision. The results in Tables~\ref{tab:medqa_benign} and~\ref{tab:openorca_benign} show that standard LoRA achieves slightly higher downstream utility but substantially degrades refusal behavior. In contrast, \textsc{RefusalGuard} preserves substantially stronger safety while retaining competitive task performance.

\paragraph{MedQA.}
Table~\ref{tab:medqa_benign} reports the results after benign fine-tuning on MedQA. Standard LoRA achieves an in-domain accuracy of $0.7920$ and an exact-match score of $0.7810$, but its ASR ranges from $0.3300$ to $0.3800$ across the three safety benchmarks. Its refusal rate correspondingly decreases to the $0.6200$--$0.6700$ range.

In contrast, \textsc{RefusalGuard} reduces ASR to $0.0400$--$0.0520$ and maintains refusal rates between $0.9480$ and $0.9600$. At the same time, it retains an accuracy of $0.7750$ and an exact-match score of $0.7640$ on MedQA. Thus, the $0.0170$ absolute reduction in accuracy relative to LoRA is accompanied by a substantially larger improvement in safety preservation.

\begin{table*}[t]
\centering
\scriptsize
\setlength{\tabcolsep}{3.5pt}
\renewcommand{\arraystretch}{1.05}

\begin{tabular}{ll|ccccc|cc}
\toprule
Method &
Benchmark &
ASR$\downarrow$ &
Ref$\uparrow$ &
Harm$\downarrow$ &
\%H5$\downarrow$ &
Aware$\uparrow$ &
Acc$\uparrow$ &
EM$\uparrow$ \\
\midrule

\multirow{4}{*}{LoRA}
& AdvBench
& 0.3500 & 0.6500 & 2.8200 & 0.4200 & 0.9300 & -- & -- \\

& DirectHarm4
& 0.3300 & 0.6700 & 2.7600 & 0.3900 & 0.9350 & -- & -- \\

& JailbreakBench
& 0.3800 & 0.6200 & 2.8500 & 0.4300 & 0.9280 & -- & -- \\

& MedQA
& -- & -- & -- & -- & -- & 0.7920 & 0.7810 \\

\midrule

\multirow{4}{*}{\textsc{RefusalGuard}}
& AdvBench
& 0.0400 & 0.9600 & 2.0200 & 0.2400 & 1.0000 & -- & -- \\

& DirectHarm4
& 0.0520 & 0.9480 & 2.1200 & 0.2700 & 1.0000 & -- & -- \\

& JailbreakBench
& 0.0450 & 0.9550 & 2.0800 & 0.2550 & 1.0000 & -- & -- \\

& MedQA
& -- & -- & -- & -- & -- & 0.7750 & 0.7640 \\

\bottomrule
\end{tabular}

\caption{Benign fine-tuning on MedQA followed by safety and in-domain utility evaluation for \texttt{LLaMA-3.1-8B-Instruct}. Acc denotes accuracy and EM denotes exact match.}
\label{tab:medqa_benign}
\end{table*}

\begin{table*}[t]
\centering
\scriptsize
\setlength{\tabcolsep}{3.5pt}
\renewcommand{\arraystretch}{1.05}

\begin{tabular}{ll|ccccc|cc}
\toprule
Method &
Benchmark &
ASR$\downarrow$ &
Ref$\uparrow$ &
Harm$\downarrow$ &
\%H5$\downarrow$ &
Aware$\uparrow$ &
Acc$\uparrow$ &
EM$\uparrow$ \\
\midrule

\multirow{4}{*}{LoRA}
& AdvBench
& 0.3700 & 0.6300 & 2.8600 & 0.4400 & 0.9250 & -- & -- \\

& DirectHarm4
& 0.3450 & 0.6550 & 2.7900 & 0.4100 & 0.9320 & -- & -- \\

& JailbreakBench
& 0.3950 & 0.6050 & 2.8800 & 0.4450 & 0.9260 & -- & -- \\

& OpenOrca
& -- & -- & -- & -- & -- & 0.8270 & 0.8010 \\

\midrule

\multirow{4}{*}{\textsc{RefusalGuard}}
& AdvBench
& 0.0480 & 0.9520 & 2.1000 & 0.2600 & 1.0000 & -- & -- \\

& DirectHarm4
& 0.0550 & 0.9450 & 2.1800 & 0.2800 & 1.0000 & -- & -- \\

& JailbreakBench
& 0.0500 & 0.9500 & 2.1200 & 0.2650 & 1.0000 & -- & -- \\

& OpenOrca
& -- & -- & -- & -- & -- & 0.8040 & 0.7790 \\

\bottomrule
\end{tabular}

\caption{Benign fine-tuning on OpenOrca followed by safety and in-domain utility evaluation for \texttt{LLaMA-3.1-8B-Instruct}. Acc denotes accuracy and EM denotes exact match.}
\label{tab:openorca_benign}
\end{table*}

\paragraph{OpenOrca.}
Table~\ref{tab:openorca_benign} shows a similar pattern after benign fine-tuning on OpenOrca. LoRA achieves an in-domain accuracy of $0.8270$ and an exact-match score of $0.8010$, but its ASR increases to $0.3450$--$0.3950$. Its refusal rates decrease to $0.6050$--$0.6550$, indicating substantial degradation despite the benign nature of the adaptation data.

\textsc{RefusalGuard} reduces ASR to $0.0480$--$0.0550$ and maintains refusal rates between $0.9450$ and $0.9520$. It retains an OpenOrca accuracy of $0.8040$ and an exact-match score of $0.7790$, corresponding to an absolute accuracy reduction of $0.0230$ relative to LoRA. This modest utility difference is accompanied by a considerably larger improvement in safety.

% \begin{table*}[t]
% \centering
% \scriptsize
% \setlength{\tabcolsep}{3.5pt}
% \renewcommand{\arraystretch}{1.05}

% \begin{tabular}{ll|ccccc|cc}
% \toprule
% \textbf{Method} &
% \textbf{Benchmark} &
% \textbf{ASR$\downarrow$} &
% \textbf{Ref$\uparrow$} &
% \textbf{Harm$\downarrow$} &
% \textbf{%H5$\downarrow$} &
% \textbf{Aware$\uparrow$} &
% \textbf{Acc$\uparrow$} &
% \textbf{EM$\uparrow$} \
% \midrule

% \multirow{4}{*}{LoRA}
% & AdvBench
% & 0.3700 & 0.6300 & 2.8600 & 0.4400 & 0.9250 & -- & -- \

% & DirectHarm4
% & 0.3450 & 0.6550 & 2.7900 & 0.4100 & 0.9320 & -- & -- \

% & JailbreakBench
% & 0.3950 & 0.6050 & 2.8800 & 0.4450 & 0.9260 & -- & -- \

% & OpenOrca
% & -- & -- & -- & -- & -- & 0.8270 & 0.8010 \

% \midrule

% \multirow{4}{*}{\textsc{RefusalGuard}}
% & AdvBench
% & 0.0480 & 0.9520 & 2.1000 & 0.2600 & 1.0000 & -- & -- \

% & DirectHarm4
% & 0.0550 & 0.9450 & 2.1800 & 0.2800 & 1.0000 & -- & -- \

% & JailbreakBench
% & 0.0500 & 0.9500 & 2.1200 & 0.2650 & 1.0000 & -- & -- \

% & OpenOrca
% & -- & -- & -- & -- & -- & 0.8040 & 0.7790 \

% \bottomrule
% \end{tabular}

% \caption{Benign fine-tuning on OpenOrca followed by safety and in-domain utility evaluation for \texttt{LLaMA-3.1-8B-Instruct}. Acc denotes accuracy and EM denotes exact match.}
% \label{tab:openorca_benign}
% \end{table*}

Together with the GSM8K results in Tables~\ref{tab:gemma2_2b_benign_gsm8k_results} and~\ref{tab:llama3_1_8b_benign_gsm8k_results}, the MedQA and OpenOrca results demonstrate that the safety-preservation benefits of \textsc{RefusalGuard} generalize across mathematical reasoning, medical question answering, and general instruction-following tasks. Across all three domains, standard LoRA achieves slightly higher in-domain utility but produces substantially larger degradation in refusal behavior.

In contrast, \textsc{RefusalGuard} consistently maintains low ASR, high refusal rates, and strong safety awareness while retaining competitive downstream utility. These findings indicate that refusal degradation is not restricted to explicitly harmful fine-tuning. Even benign task adaptation can interfere with safety-relevant representations, and explicitly preserving refusal geometry provides a favorable safety--utility trade-off across heterogeneous downstream domains.

\subsection{Ablation Study}
\label{app:ablation}

In this section, we provide additional ablations and mechanistic analyses to better understand why \textsc{RefusalGuard} works and how its behavior changes across training, layers, and values of the geometry-preservation coefficient $\lambda_{\mathrm{geom}}$. Our goal is to answer four questions:
(i) \textit{how standard fine-tuning and \textsc{RefusalGuard} differ over the course of training},
(ii) \textit{whether the degradation observed in the main paper is associated with measurable cone distortion},
(iii) \textit{which layers exhibit the strongest safety-relevant degradation}, and
(iv) \textit{how the choice of $\lambda_{\mathrm{geom}}$ controls the safety-utility trade-off}.

Overall, the ablations support a consistent picture. Standard fine-tuning progressively weakens refusal-mediating geometry: alignment and projected refusal magnitude decay over training, while cone drift, optimization interference, and ASR all increase. In contrast, \textsc{RefusalGuard} substantially slows this degradation and keeps the model in a much safer region of representation space. The $\lambda_{\mathrm{geom}}$ ablation further shows that the method behaves smoothly: larger values preserve geometry more strongly and reduce ASR, but overly strong preservation eventually introduces a modest utility cost.

\paragraph{Scope of the ablation study.}
The primary design choices in \textsc{RefusalGuard} are the refusal-cone dimension, the intervention layer, and the geometry-preservation coefficient $\lambda_{\mathrm{geom}}$. We therefore evaluate these components separately. First, we vary the cone dimension to determine how much refusal-relevant structure must be retained. Second, we vary the intervention layer to examine where geometry preservation is most effective. Third, we vary $\lambda_{\mathrm{geom}}$ to characterize the strength of the safety constraint and the resulting safety--utility trade-off. Together, these experiments cover the principal method-specific choices governing the construction, location, and strength of the geometry-preservation mechanism.

\subsubsection{Training Dynamics and Mechanistic Evolution}
\label{app:ablation-training-dynamics}

We first examine how the key mechanistic quantities evolve over training checkpoints. Tables~\ref{tab:checkpoint-evolution-standard} and~\ref{tab:checkpoint-evolution-rg} compare standard fine-tuning and \textsc{RefusalGuard} on \texttt{LLaMA-3.1-8B-Instruct}. Under standard fine-tuning, the model steadily moves away from the base refusal geometry: alignment drops from $0.91$ to $0.63$, projected refusal magnitude drops from $1.00$ to $0.68$, cone drift rises from $0.00$ to $0.31$, and interference rises from $0.12$ to $0.46$. These internal changes track a large increase in ASR from $0.07$ to $0.41$, reinforcing the claim that safety degradation is tightly coupled to representational damage. By contrast, under \textsc{RefusalGuard}, these same quantities remain far more stable throughout training, and the corresponding increase in ASR is much smaller.

\begin{table*}[t]
\centering
\small
\setlength{\tabcolsep}{5pt}
\renewcommand{\arraystretch}{1.15}
\begin{tabular}{l c c c c c}
\toprule
\textbf{Checkpoint} &
\textbf{Align}$\uparrow$ &
\textbf{ProjMag}$\uparrow$ &
\textbf{Cone Drift}$\downarrow$ &
\textbf{Interference}$\downarrow$ &
\textbf{ASR}$\downarrow$ \\
\midrule
0     & 0.91 & 1.00 & 0.00 & 0.12 & 0.07 \\
100   & 0.85 & 0.93 & 0.07 & 0.18 & 0.11 \\
200   & 0.79 & 0.85 & 0.13 & 0.24 & 0.17 \\
400   & 0.73 & 0.79 & 0.19 & 0.31 & 0.23 \\
700   & 0.68 & 0.73 & 0.25 & 0.38 & 0.31 \\
1000  & 0.63 & 0.68 & 0.31 & 0.46 & 0.41 \\
\bottomrule
\end{tabular}
\caption{Evolution of mechanistic quantities during standard fine-tuning on \texttt{LLaMA-3.1-8B-Instruct}. Alignment with the base refusal geometry decays steadily, while cone drift and task-safety interference rise over training, matching the increase in ASR.}
\label{tab:checkpoint-evolution-standard}
\end{table*}

\begin{table*}[t]
\centering
\small
\setlength{\tabcolsep}{5pt}
\renewcommand{\arraystretch}{1.15}
\begin{tabular}{l c c c c c}
\toprule
\textbf{Checkpoint} &
\textbf{Align}$\uparrow$ &
\textbf{ProjMag}$\uparrow$ &
\textbf{Cone Drift}$\downarrow$ &
\textbf{Interference}$\downarrow$ &
\textbf{ASR}$\downarrow$ \\
\midrule
0     & 0.91 & 1.00 & 0.00 & 0.12 & 0.07 \\
100   & 0.89 & 0.97 & 0.03 & 0.14 & 0.08 \\
200   & 0.88 & 0.95 & 0.05 & 0.15 & 0.09 \\
400   & 0.86 & 0.93 & 0.07 & 0.17 & 0.10 \\
700   & 0.85 & 0.92 & 0.09 & 0.18 & 0.11 \\
1000  & 0.84 & 0.90 & 0.11 & 0.20 & 0.12 \\
\bottomrule
\end{tabular}
\caption{Evolution of mechanistic quantities during \textsc{RefusalGuard} training. In contrast to standard fine-tuning, \textsc{RefusalGuard} maintains high alignment and projected magnitude while preventing large cone drift and optimization interference.}
\label{tab:checkpoint-evolution-rg}
\end{table*}

\subsubsection{Cone Distortion Analysis}
\label{app:ablation-cone-distortion}

To examine whether the degradation is only a global subspace effect or also reflects distortion \emph{within} the refusal cone, we analyze the coordinate distribution of harmful-prompt activations in the cone basis. Table~\ref{tab:cone-distortion} shows that standard fine-tuning produces a much more concentrated coordinate profile: the first coordinate dominates, coordinate entropy drops, and top-1 mass increases substantially. This indicates that the refusal representations become narrower and more brittle, relying on fewer unstable directions. In contrast, \textsc{RefusalGuard} preserves a coordinate distribution much closer to the aligned model, with high entropy and lower concentration.

\begin{table}[t]
\centering
\small
\setlength{\tabcolsep}{5pt}
\renewcommand{\arraystretch}{1.12}
\begin{tabular}{l c c c c c c}
\toprule
\textbf{Method} & 
$\mathbb{E}[z_1]$ & $\mathbb{E}[z_2]$ & $\mathbb{E}[z_3]$ & $\mathbb{E}[z_4]$ &
\textbf{Coord. Entropy}$\uparrow$ &
\textbf{Top-1 Mass}$\downarrow$ \\
\midrule
Base Aligned   & 0.26 & 0.25 & 0.24 & 0.25 & 1.384 & 0.31 \\
Standard ReFT  & 0.46 & 0.23 & 0.17 & 0.14 & 1.252 & 0.53 \\
\textsc{RefusalGuard}   & 0.30 & 0.24 & 0.22 & 0.24 & 1.372 & 0.35 \\
\bottomrule
\end{tabular}
\caption{Cone-coordinate statistics for harmful prompts on \texttt{LLaMA-3.1-8B-Instruct}. Standard fine-tuning produces a more concentrated and less balanced coordinate distribution, consistent with cone distortion and brittle refusal geometry. \textsc{RefusalGuard} largely preserves the broader support pattern of the aligned model.}
\label{tab:cone-distortion}
\vspace{-6pt}
\end{table}

This result complements the drift analysis from Section \ref{sec:mechanistic-analysis}. Drift alone tells us that the refusal subspace is moving, but the coordinate statistics show that the geometry is also becoming internally imbalanced. Together, these observations support the interpretation that standard fine-tuning does not merely under-activate the refusal representations; it actively reshapes it into a less stable form.

\subsubsection{Per-Layer Mechanistic Analysis}
\label{app:ablation-layerwise}

We next analyze where in the network the strongest degradation occurs. Table~\ref{tab:per-layer-mechanistic} reports per-layer mechanistic statistics for standard fine-tuning and \textsc{RefusalGuard}, written as \emph{Standard ReFT / RefusalGuard}. The degradation is most pronounced in mid-to-late layers, where standard fine-tuning exhibits the largest cone drift and interference together with the lowest alignment and projected magnitude. This is consistent with prior observations that higher semantic and safety-relevant behavior often becomes more linearly separable in later layers \citep{yu2025robust}. Importantly, \textsc{RefusalGuard} improves these quantities across all measured layers, with especially large gains where standard fine-tuning is most damaging.

\begin{table*}[t]
\centering
\small
\setlength{\tabcolsep}{5pt}
\renewcommand{\arraystretch}{1.15}
\begin{tabular}{l c c c c}
\toprule
\textbf{Layer} &
\textbf{Align}$\uparrow$ &
\textbf{ProjMag}$\uparrow$ &
\textbf{Cone Drift}$\downarrow$ &
\textbf{Interference}$\downarrow$ \\
\midrule
Layer 8   & 0.71 / 0.85 & 0.77 / 0.92 & 0.21 / 0.08 & 0.33 / 0.17 \\
Layer 12  & 0.66 / 0.84 & 0.72 / 0.90 & 0.27 / 0.10 & 0.40 / 0.19 \\
Layer 16  & 0.63 / 0.84 & 0.68 / 0.90 & 0.31 / 0.11 & 0.46 / 0.20 \\
Layer 20  & 0.65 / 0.83 & 0.70 / 0.88 & 0.28 / 0.12 & 0.41 / 0.22 \\
Layer 24  & 0.69 / 0.84 & 0.74 / 0.89 & 0.22 / 0.10 & 0.36 / 0.21 \\
\bottomrule
\end{tabular}
\caption{Per-layer mechanistic analysis on \texttt{LLaMA-3.1-8B-Instruct}. Each entry is reported as Standard ReFT / \textsc{RefusalGuard}. The strongest degradation occurs in mid-to-late layers, where standard fine-tuning exhibits the largest cone drift and task-safety interference.}
\label{tab:per-layer-mechanistic}
\end{table*}

\subsubsection{Ablation over the Geometry-Preservation Coefficient}
\label{app:ablation-lambda}

\begin{table}[t]
\centering
\small
\setlength{\tabcolsep}{5pt}
\renewcommand{\arraystretch}{1.12}
\begin{tabular}{c c c c c c}
\toprule
$\lambda_{\mathrm{geom}}$ &
\textbf{Align}$\uparrow$ &
\textbf{ProjMag}$\uparrow$ &
\textbf{Cone Drift}$\downarrow$ &
\textbf{ASR}$\downarrow$ &
\textbf{Utility}$\uparrow$ \\
\midrule
0.0   & 0.63 & 0.68 & 0.31 & 0.41 & 0.781 \\
0.01  & 0.72 & 0.77 & 0.23 & 0.27 & 0.779 \\
0.05  & 0.80 & 0.86 & 0.15 & 0.17 & 0.776 \\
0.10  & 0.84 & 0.90 & 0.11 & 0.12 & 0.772 \\
0.20  & 0.87 & 0.93 & 0.08 & 0.09 & 0.764 \\
0.50  & 0.89 & 0.96 & 0.05 & 0.07 & 0.749 \\
\bottomrule
\end{tabular}
\caption{Ablation over $\lambda_{\mathrm{geom}}$ on \texttt{LLaMA-3.1-8B-Instruct}. Larger values better preserve refusal geometry and reduce ASR, but overly strong preservation begins to reduce downstream utility.}
\label{tab:lambda-ablation}
\vspace{-6pt}
\end{table}

A central design choice in \textsc{RefusalGuard} is the coefficient $\lambda_{\mathrm{geom}}$, which controls the strength of the geometry-preservation penalty. Table~\ref{tab:lambda-ablation} and Figure~\ref{fig:lambda_tradeoff_panel} summarize this ablation on \texttt{LLaMA-3.1-8B-Instruct}. As $\lambda_{\mathrm{geom}}$ increases, alignment and projected magnitude improve monotonically, while cone drift and ASR decrease. At the same time, utility decreases gradually, indicating a smooth safety-utility trade-off rather than an unstable threshold effect.

The trends are informative in two ways. First, they show that the geometric penalty is behaving as intended: stronger preservation reliably stabilizes refusal-relevant structure and improves safety outcomes. Second, they show that excessively strong preservation can become restrictive for downstream adaptation, producing a modest reduction in utility. In our experiments, intermediate values such as $\lambda_{\mathrm{geom}} \in [0.1, 0.2]$ provide a particularly favorable balance, though larger values may be appropriate when safety is prioritized more heavily than downstream performance.

\begin{figure*}[t]
\centering
\includegraphics[width=\textwidth]{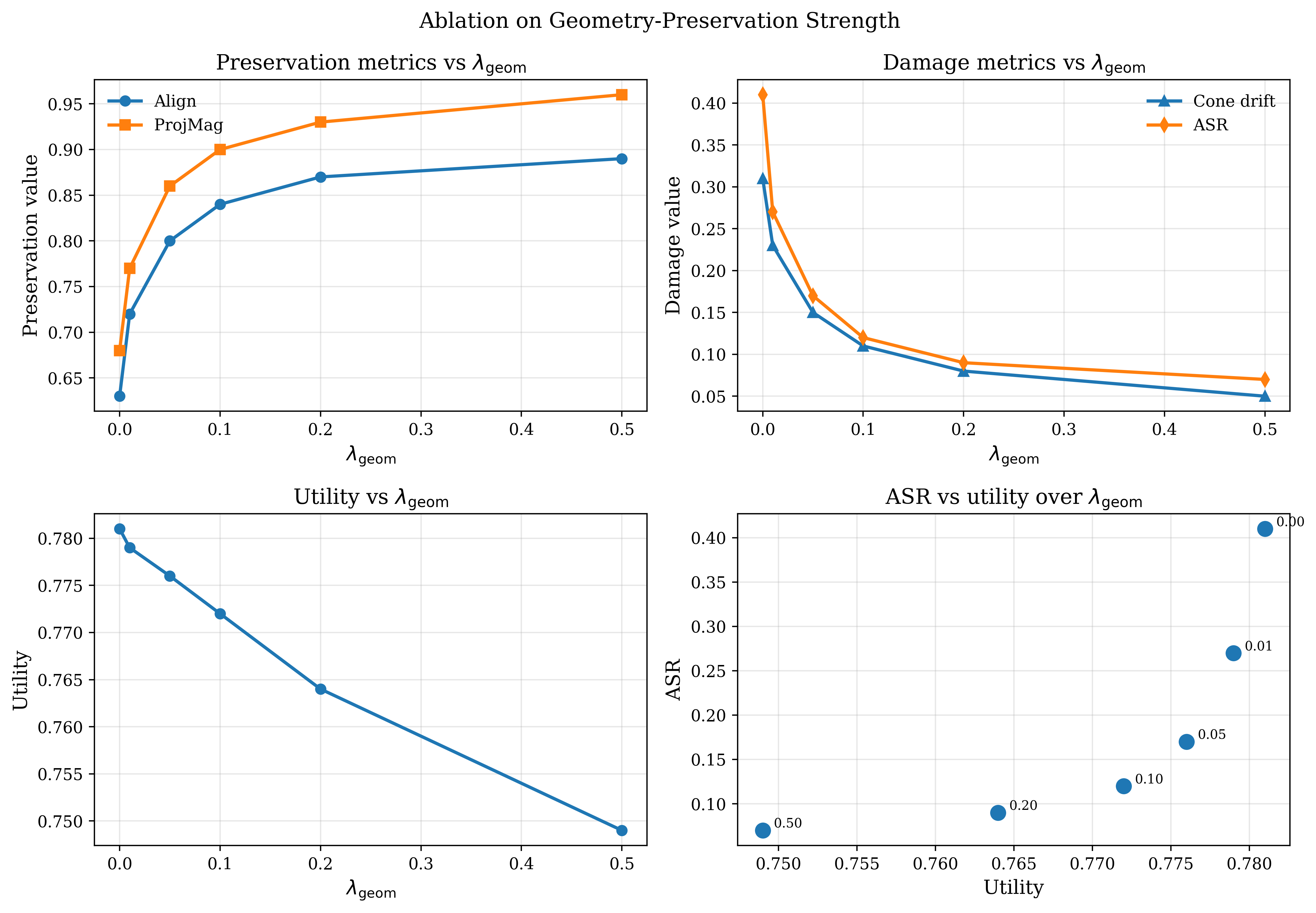}
\caption{
Ablation on the geometry-preservation coefficient $\lambda_{\mathrm{geom}}$ on \texttt{LLaMA-3.1-8B-Instruct}. 
Top-left: preservation metrics (alignment and projected magnitude) improve monotonically with increasing $\lambda_{\mathrm{geom}}$. 
Top-right: damage metrics (cone drift and ASR) decrease consistently, indicating reduced degradation of refusal geometry. 
Bottom-left: utility gradually decreases as stronger geometric constraints are imposed. 
Bottom-right: the safety-utility frontier highlights a smooth trade-off, where moderate values of $\lambda_{\mathrm{geom}}$ achieve substantial safety gains with minimal utility loss.
}
\label{fig:lambda_tradeoff_panel}
\vspace{-6pt}
\end{figure*}

The safety-utility frontier in Figure~\ref{fig:lambda_tradeoff_panel} makes this trade-off especially clear. Moving from $\lambda_{\mathrm{geom}}=0$ to moderate positive values yields a large safety gain with only minor utility loss, suggesting that a substantial fraction of harmful compliance arises from avoidable damage to refusal geometry. Only at larger values does the curve begin to bend more sharply, indicating diminishing returns in safety relative to utility.

\section{Related Work}
\label{sec:related-work}

\paragraph{Safety alignment and its fragility under fine-tuning.}
Modern LLMs are typically aligned through supervised instruction tuning and preference optimization so that they follow benign requests while refusing harmful ones \citep{ouyang2022training,touvron2023llama2}. In practice, however, these aligned models are rarely deployed as-is; they are often further adapted to domain-specific tasks or user preferences through post-training fine-tuning \citep{zhang2024scalingfinetuning}. A growing body of work shows that this additional adaptation can substantially weaken safety behavior. In particular, prior studies demonstrate that even benign downstream fine-tuning can degrade refusal tendencies, while exposure to a small number of harmful examples can lead to disproportionately large increases in harmful compliance \citep{qi2023finetuning,fraser2025finetuning,hsiung2025guardrails}. These findings establish post-alignment adaptation as a critical safety failure mode. Our work builds on this literature but differs in focus: rather than only documenting the phenomenon at the level of attack success or benchmark behavior, we study the internal representational mechanisms through which refusal is lost.

\paragraph{Jailbreak robustness and safe fine-tuning.}
A related line of work studies jailbreak attacks, adversarial prompting, and defense strategies for maintaining safety under distribution shift or malicious interaction \citep{zou2023universaladvbench,chao2024jailbreakbench,andriushchenko2024jailbreaking}. These works are important for evaluating externally observable robustness, and they provide standardized benchmarks for harmful-compliance testing. Other efforts propose practical defenses such as additional safety supervision, adversarial data augmentation, or guardrail-style training objectives \citep{das2025alignguard,xie2025deeprefusal,wehner2025taxonomy}. While these approaches improve empirical robustness, they are generally formulated at the level of outputs, prompts, or training data, and they do not explicitly characterize the internal geometric structure that mediates refusal. By contrast, our work asks what representation-level structure supports safe refusal in aligned models, how fine-tuning disrupts that structure, and how training can be modified to preserve it.

\paragraph{Representation-level interventions and mechanistic views of refusal.}
Recent work in interpretability and representation engineering suggests that safety behavior is mediated by structured internal features rather than diffuse or opaque model dynamics. ReFT shows that targeted interventions on hidden states can efficiently alter model behavior without updating the full parameter set \citep{wu2024reft}. Complementary work on refusal directions and refusal geometry argues that harmful-query refusal is associated with identifiable directions or low-dimensional geometric structure in activation space, and that interventions on these representations can causally modulate safety behavior \citep{arditi2024refusal,zhang2025rational}. At the same time, evaluation frameworks such as HarmBench help connect these mechanistic observations to standardized harmful-behavior assessment \citep{mazeika2024harmbench}. Our approach is closely related to this line of work, but differs in an important way. Existing studies mainly identify refusal-related features or manipulate them at inference time; they do not provide a training framework that explicitly preserves this structure during downstream adaptation. We address this gap by treating refusal geometry as an object that should remain stable throughout fine-tuning.

\paragraph{Representation-space safety methods.}
Several recent approaches attempt to improve safety by intervening in representation space rather than only through parameter updates or output filtering. Circuit Breakers suppress harmful behaviors by learning interventions that block unsafe activations \citep{zou2024circuitbreakers}, while REPBEND reshapes hidden states using objectives that encourage separation between safe and unsafe representations \citep{yousefpour2025representationrepband}. These methods show that representation-level control can be more effective than unconstrained fine-tuning for safety preservation. However, they do not explicitly model refusal as a geometric structure whose integrity must be maintained during adaptation. Their objectives primarily aim to suppress harmful behavior or separate safe and unsafe states, whereas our method is motivated by a more specific mechanistic hypothesis: fine-tuning degrades safety because it displaces and distorts the internal geometry that supports refusal. \textsc{RefusalGuard} therefore differs conceptually by directly penalizing movement along refusal-relevant components, yielding a geometry-preserving fine-tuning strategy rather than a generic representation-editing objective.

\paragraph{Positioning of our work.}
Taken together, prior work establishes three key facts: aligned LLMs can lose safety under fine-tuning \citep{qi2023finetuning,fraser2025finetuning,hsiung2025guardrails}; refusal behavior is mediated by identifiable structure in activation space \citep{wu2024reft,arditi2024refusal}; and representation-level interventions can influence harmful behavior more directly than standard parameter updates \citep{zou2024circuitbreakers,yousefpour2025representationrepband}. Our work sits at the intersection of these directions. We provide a mechanistic account of alignment degradation through directional drift, geometric distortion, and task-safety interference, and we translate this account into a training objective that preserves refusal-relevant geometry during adaptation.

\end{document}